\pdfoutput=1

\documentclass[11pt]{article}

\usepackage{ACL2023}

\usepackage{times}
\usepackage{latexsym}
\usepackage{arydshln}
\usepackage{graphicx}
\usepackage{subcaption}
\usepackage{hyperref}
\usepackage{booktabs,arydshln}
\usepackage{amsmath}
\usepackage{enumitem}
\usepackage{multirow}
\usepackage[export]{adjustbox}
\usepackage[utf8]{inputenc}

\makeatletter
\def\adl@drawiv#1#2#3{%
        \hskip.5\tabcolsep
        \xleaders#3{#2.5\@tempdimb #1{1}#2.5\@tempdimb}%
                #2\z@ plus1fil minus1fil\relax
        \hskip.5\tabcolsep}
\newcommand{\cdashlinelr}[1]{%
  \noalign{\vskip\aboverulesep
           \global\let\@dashdrawstore\adl@draw
           \global\let\adl@draw\adl@drawiv}
  \cdashline{#1}
  \noalign{\global\let\adl@draw\@dashdrawstore
           \vskip\belowrulesep}}
\makeatother

\usepackage[T1]{fontenc}

\usepackage[utf8]{inputenc}

\usepackage{microtype}

\usepackage{inconsolata}

\title{Multilingual Event Extraction from Historical Newspaper Adverts \\
\normalsize \textcolor{orange}{WARNING: This paper shows dataset samples which are racist in nature}}

\author{Nadav Borenstein \\
  University of Copenhagen \\
  \texttt{nadav.borenstein@di.ku.dk} \\\And
  Natália da Silva Perez \\
  Erasmus University Rotterdam \\
  \texttt{dasilvaperez@eshcc.eur.nl} \\
  \AND
  Isabelle Augenstein \\
  University of Copenhagen \\
  \texttt{augenstein@di.ku.dk} \\}

\begin{document}
\maketitle
\begin{abstract}
NLP methods can aid historians in analyzing textual materials in greater volumes than manually feasible. Developing such methods poses substantial challenges though. First, acquiring large, annotated historical datasets is difficult, as only domain experts can reliably label them. Second, most available off-the-shelf NLP models are trained on modern language texts, rendering them significantly less effective when applied to historical corpora. This is particularly problematic for less well studied tasks, and for languages other than English. 
This paper addresses these challenges while focusing on the under-explored task of event extraction from a novel domain of historical texts. We introduce a new multilingual dataset in English, French, and Dutch composed of newspaper ads from the early modern colonial period reporting on enslaved people who liberated themselves from enslavement. We find that: 1) even with scarce annotated data, it is possible to achieve surprisingly good results by formulating the problem as an extractive QA task and leveraging existing datasets and models for modern languages; and 2) cross-lingual low-resource learning for historical languages is highly challenging, and machine translation of the historical datasets to the considered target languages is, in practice, often the best-performing solution.

\end{abstract}

\section{Introduction}
\label{sec:intro}
Analyzing large corpora of historical documents can provide invaluable insights on past events in multiple resolutions, from the life of an individual to processes on a global scale \cite{nadav2023karolina, laite2020emmet, gerritsen2012scales}. While historians traditionally work closely with the texts they study, automating parts of the analysis using NLP tools can help speed up the research process and facilitate the extraction of historical evidence from large corpora, allowing historians to focus on interpretation.
    
However, building NLP models for historical texts poses a substantial challenge. First, acquiring large, annotated historical datasets is difficult \cite{hamalainen-etal-2021-lemmatization, bollmann-sogaard-2016-improving}, as only domain experts can reliably label them. This renders the default fully-supervised learning setting less feasible for historical corpora. Compounding this, most off-the-shelf NLP models were trained on modern language texts and display significantly weaker performance for historical documents \cite{jdmdh:9690, baptiste2021transferring, hardmeier-2016-neural}, which usually suffer from a high rate of OCR errors and are written in a substantially different language. This is particularly challenging for less well-studied tasks or for non-English languages.

One of these under-explored tasks is event extraction from historical texts \cite{10.1162/coli_a_00347, historical_event_extraction_2021}, which can aid in retrieving information about complex events from vast amounts of texts. Here, we research extraction of events from adverts in colonial newspapers reporting on enslaved people who escaped their enslavers. Studying these ads can shed light on the linguistic processes of racialization during the early modern colonial period (c. 1450 to 1850), the era of the transatlantic slave trade, which coincided with the early era of mass print media. 

Methodologically, we research low-resource learning methods for event extraction, for which only a handful of prior papers exist \cite{historical_event_extraction_2021, 10.1162/coli_a_00347}. To the best of our knowledge, this is the first paper to study historical event extraction in a multilingual setting.

\begin{figure*}[t]
    \centering
         \includegraphics[width=\textwidth, trim={0 10cm 0 0},clip]{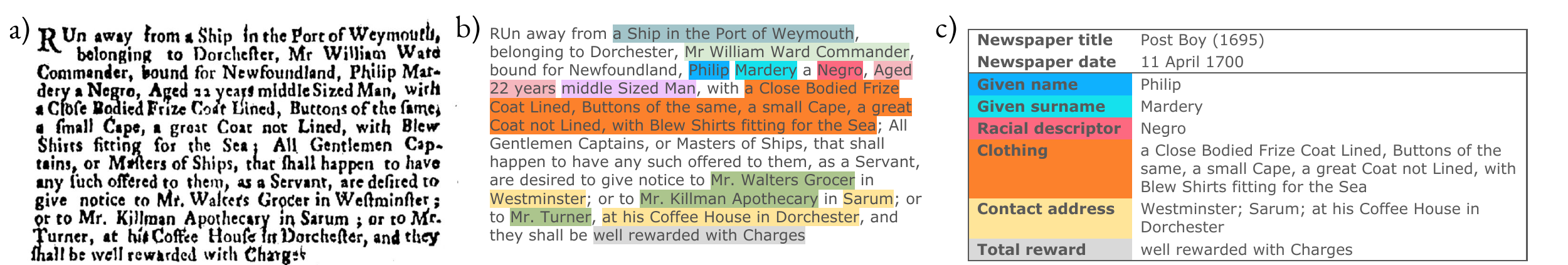}
         \caption{An example from the annotated \textit{Runaway Slaves in Britain} dataset. Each data point includes a scan of the ad (a), the extracted text (b), and a list of attributes that appear in the ad as well as relevant metadata (c).}
         \label{fig:runaways_example}
\end{figure*}

Specifically, our contributions are as follows:
\begin{itemize}[noitemsep]
    \item We construct a new multilingual dataset in English, French, and Dutch of \textit{``freedom-seeking events''}, composed of ads placed by enslavers reporting on enslaved people who sought freedom by escaping them, building on an existing annotated English language dataset of ``runaway slave adverts'' \cite{simon_p_newman_runaway_nodate}.\footnote{We make our dataset and code publicly available at \url{https://github.com/nadavborenstein/EE-from-historical-ads}} Fig. \ref{fig:runaways_example}a contains an example ad.
    \item We propose to frame event extraction from historical texts as extractive question answering. We show that even with scarce annotated data, this formulation can achieve surprisingly good results by leveraging existing resources for modern languages.
    \item We show that cross-lingual low-resource learning for historical languages is highly challenging, and machine translation of the historical datasets to the target languages is often the best-performing solution in practice.
\end{itemize}

\section{Related Work}
\label{sec:related_work}
\subsection{NLP for Historical Texts}

Prior work on NLP for historical texts has mainly focused on OCR and text normalization \cite{spell_correction_2017, robertson-goldwater-2018-evaluating, text_normalization_2018b, bollmann-2019-large, spell_correction_2021}. However, NLP has also been used to assist historians in analyzing large amounts of textual material in more complex ways. Recent work has researched tasks such as PoS tagging \cite{YangE16},
Named Entity Recognition \cite{NER_2021,de-toni-etal-2022-entities} and co-reference resolution \cite{coreference_2022, coreference_2015}, and bias analysis \cite{nadav2023karolina}. Many of these studies report the difficulties of acquiring large annotated historical datasets \cite{hamalainen-etal-2021-lemmatization, bollmann-sogaard-2016-improving} and replicating the impressive results of large pre-trained language models on modern texts \cite{historical_event_extraction_2021, de-toni-etal-2022-entities}. This also led prior work to focus on monolingual texts, particularly in English, while neglecting low-resource languages.  
In this paper, we attempt to alleviate these challenges while investigating a task that is underexplored from the perspective of historical NLP -- multilingual event extraction.

\subsection{Event Extraction}

Event extraction \cite{event_extraction_survey_2011, event_extraction_survey_2018} is the task of organising natural text into structured events -- specific occurrences of something that happens at a particular time and place involving one or more participants, each associated with a set of attributes. 

Traditionally, event extraction is decomposed into smaller, less complex subtasks \cite{lin-etal-2020-joint, li-etal-2020-event}, such as detecting the existence of an event \cite{event_detection_2011, event_detection_2018, event_detection_2019}, identifying its participants \cite{event_participants_2021, li-etal-2020-event}, and extracting the attributes associated with the event \cite{li-etal-2020-event, event_arguments_2020, du-cardie-2020-event}. Recent work \cite{liu-etal-2020-event, du-cardie-2020-event} has shown the benefit of framing event extraction as a QA task, especially for the sub-task of attribute extraction, which is the focus of this work. We build on the latter finding, by framing the identification of attributes associated with historical events as an extractive QA task. 

Event extraction from historical texts is much less well studied than extraction from modern language texts, with only a handful of works targeting this task. \citet{historical_event_extraction_2011,historical_event_extraction_2011b} develop simple pipelines for extracting knowledge about historical events from modern Dutch texts. \citet{10.1162/coli_a_00347} define annotation guidelines for detecting and classifying events mentioned in historical texts and compare two models on a new corpus of historical documents. \citet{boros2022assessing} study the robustness of two event detection models to OCR noise by automatically degrading modern event extraction datasets in several languages. Finally, and closer to this work, \citet{historical_event_extraction_2021} present BRAD, a dataset for event extraction from English historical texts about Black rebellions, which is not yet publicly available. 
They find that there is a significant gap in the performance of current models on BRAD compared to modern datasets. Conversely, we explore event extraction in a multilingual setting while performing a more exhaustive evaluation of various models and pipelines.

\section{Methods}
\label{sec:methods}
We now describe the methodology of the paper, including problem formulation (§\ref{sec:formulation}), datasets (§\ref{sec:datasets}), models (§\ref{sec:models}), and the experiments setup (§\ref{sec:setup}).

\subsection{Problem Formulation}
\label{sec:formulation}

\begin{figure}
\centering
     \includegraphics[width=\columnwidth]{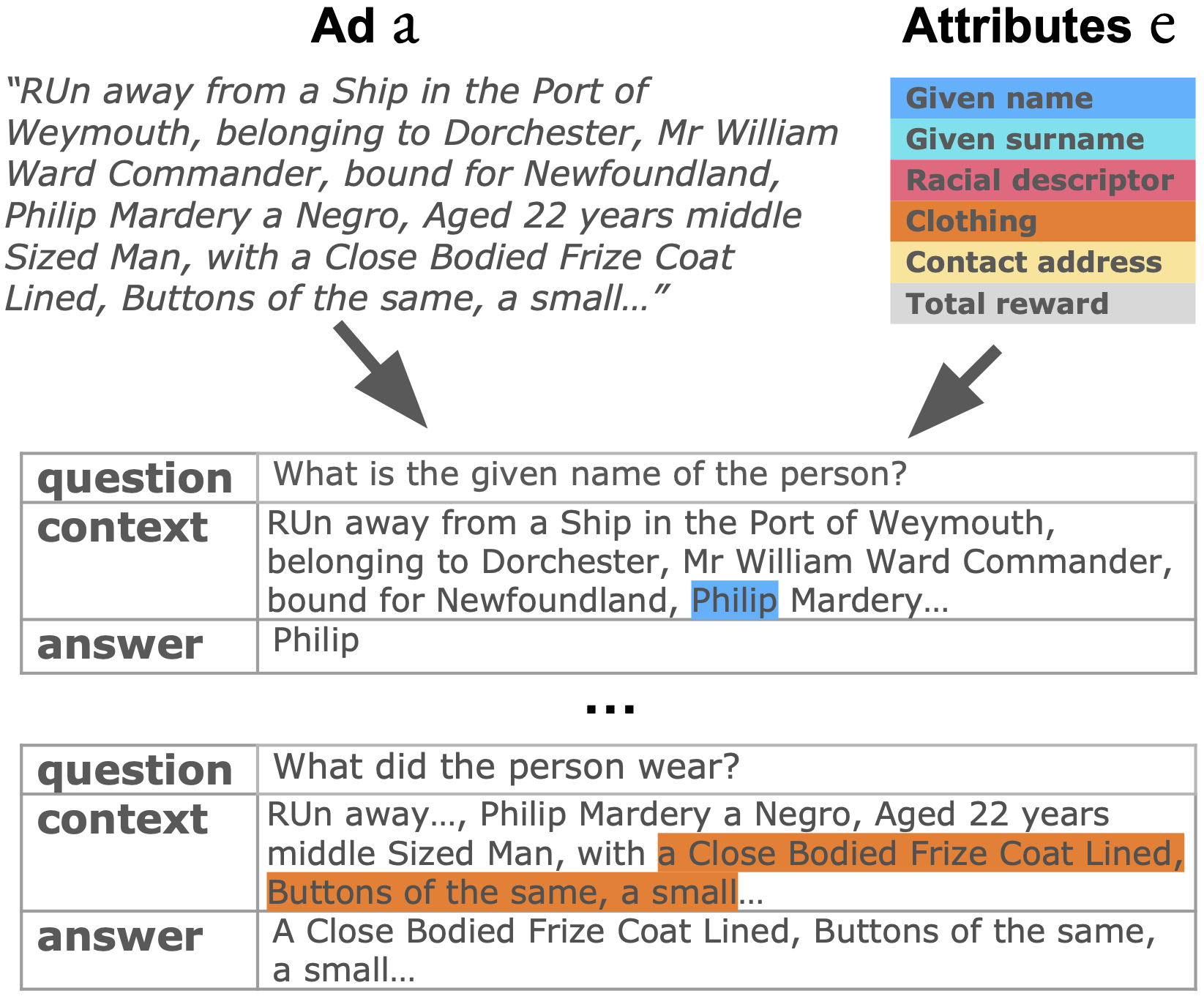}
     \caption{Our data processing pipeline: each ad is converted to a collection of extractive QA examples, where each attribute is mapped to a natural language question.}
     \label{fig:pipeline}
\end{figure}

Our starting point is a dataset where each sample is an ad corresponding to a single event. Therefore, we do not have to use event triggers – we already know what event appeared in each sample (a freedom-seeking event). We focus instead on the sub-task of attribute extraction. 
Following prior work \cite{liu-etal-2020-event}, we formulate the problem as an extractive QA task (see Fig. \ref{fig:pipeline}). Specifically, given an advert $a$ and an event attribute $e$, we convert $e$ into a natural question $q$ and search for a text span in $a$ that answers $q$. We convert the attributes to questions manually;\footnote{We assume a small number of well-defined attributes of interest, as is common for historical research.} see §\ref{sec:datasets} for details. For example, if $a$ is the attribute ``total reward'', we look for a text span in $a$ that answers the question ``How much reward is offered?''. 

We opt for this formulation for several reasons. First, extractive QA has the advantage of retrieving event attributes in the form of a span that appears verbatim in the historical document. This feature is crucial for historians, who might not trust other types of output (an abstractive QA model might generate paraphrases of the attribute or even hallucinate nonexistent facts \cite{zhou-etal-2021-detecting}). 

Second, this formulation is especially useful in low resource settings. As annotating historical corpora is expensive and labour-intensive, these settings are prevalent in historical domains. Extractive QA is a well-researched task, with many existing datasets \cite{squad, xsquad:2019, adver_qa} and model checkpoints \cite{deepset_roberta_base, deepset_xlm_roberta_base} targeting this problem. While based on modern text, the checkpoints could still be used for transfer learning (§\ref{sec:models} lists the models we use for transfer learning).    

Finally, an extractive QA formulation is efficient -- as each event is composed of different attributes, each of which becomes a single training instance, one annotated historical ad corresponds to multiple training examples. In addition, a single model can be applied to all attribute types. This allows for a simpler and cheaper deployment, as well as a model that can benefit from multitask training and can more easily generalize to unseen attributes (§\ref{sec:analysis}). 

Note that here we assume a dataset where each sample is an ad corresponding to a single self-liberation event. This setting differs from works focusing on the sub-task of event detection, e.g. using event triggers \cite{event_detection_2019}.

\subsection{Datasets}
\label{sec:datasets}

\begin{table*}[t]
\centering
\fontsize{10}{10}\selectfont
 \begin{tabular}{llrrr}
 \toprule
  Dataset & Language  & $\#$Labeled ads & $\#$Labeled
  attributes & $\#$Unlabeled ads \\ 
  \midrule 
  Runaways Slaves in Britain & en & $835$ & $8\,270$ & $0$ \\  %
  Runaways Slaves in Britain & fr (translated) & $834$ & $8\,238$ & $0$ \\
  Runaways Slaves in Britain & nl (translated) & $834$ & $8\,234$ & $0$ \\ \midrule
  Marronage & en  & $0$ & $0$ & $3\,026$ \\ 
  Marronage & fr & $41$ & $313$ & $19\,066$ \\ \midrule
  Delpher & nl & $44$ & $272$ & $2\,742$\space \space issues \\ \bottomrule
  
\end{tabular}
 \caption{Sizes of the different datasets.}
  \label{tab:datasets}
\end{table*}

We use a combination of annotated and unannotated datasets in three languages from different sources. See Tab. \ref{tab:datasets} for a summary of the datasets and their respective sizes. 

    \textbf{Annotated Dataset} The primary resource we use in our evaluation is an annotated English dataset scraped from the website of the \textit{Runaways Slaves in Britain} project \cite{simon_p_newman_runaway_nodate}, a searchable database of over $800$ newspaper adverts printed between 1700 and 1780 placed by enslavers who wanted to capture enslaved people who had self-liberated. Each ad was manually transcribed and annotated with more than $50$ different attributes, such as the described gender and age, what clothes the enslaved person wore, and their physical description. See Fig. \ref{fig:runaways_example} for an example instance.  

    We clean and split the dataset into training and validation sets (70 / 30\% split), and pre-process it to match the format of SQuAD-v2 \cite{squad}, a large benchmark for extractive QA.\footnote{We had to discard some attributes and annotations as the annotations did not always appear verbatim in the adverts and, in some cases, could not be mapped back to the ads.} This involves converting each attribute into a natural language question. To find the best natural question for each attribute we first manually generate five natural questions per attribute. 
    We then take a frozen pre-trained extractive QA model (RoBERTa-base \cite{liu2019roberta} fine-tuned on SQuAD-v2) and use it to predict that attribute from the train set using each candidate question. We choose the question that results in the highest SQuAD-v2 $F1$ \cite{rajpurkar-etal-2018-know}.
    Tab. \ref{tab:attributes} in App. \ref{app:attributes} lists the resulting attributes paired with natural questions. 
    
    As no comparable datasets exist for languages other than English, we automatically translated the training split of the \textit{Runaway Slaves in Britain} dataset into French and Dutch to support supervised training in those languages. To ensure the quality of the translation, we asked native speakers to rate 20 translations on a Likert scale of 1-5 for accuracy and fluency. 
    Tab. \ref{tab:translation_evaluation_human} in App. \ref{App:translation} suggests that the quality of the translations is sufficiently good. 
    However, the translation process may have introduced a bias towards modern language, which could affect performance on these languages compared to English (§\ref{sec:results}). See App. \ref{App:translation} for a description of the translation process and its evaluation.

    \textbf{Unannotated datasets} In addition to the relatively small annotated dataset in English, we also collected an unannotated dataset of adverts in French and English scraped from \textit{Marronage dans le monde atlantique},\footnote{\url{www.marronnage.info/fr/index.html}} a platform that contains more than 20,000 manually transcribed newspaper ads about escaped enslaved people, published in French and English between the years 1765 -- 1833. 
    
    For Dutch, no datasets of pre-extracted ads of such events exist yet, and we thus manually construct it. 
    We use 2,742 full issues of the newspaper \textit{De Cura{\c c}aosche courant}, scraped from \textit{Delpher},\footnote{\url{www.delpher.nl}} a searchable API of millions of digitized OCRd texts from Dutch newspapers, books and magazines from all time periods. \textit{De Cura{\c c}aosche courant} was chosen because almost all its issues from 1816 -- 1882 are available, and it was printed mostly in Dutch (with some sections in other languages) in the Caribbean island of Cura{\c c}ao, a Dutch colony during the time period we are concerned with. It is worth noting that, due to the OCR process, this dataset is noisier than the others mentioned above.
    
    \textbf{Multilingual evaluation dataset} To accurately evaluate our methods on French and Dutch in addition to English, two historians of the early modern period who work with those languages manually annotated $41$ and $44$ adverts from the French \textit{Marronage} and the Dutch \textit{Delpher} corpora, respectively. As our Dutch dataset is composed of entire newspaper issues and not individual ads, the historians had first to find relevant ads before they could annotate them.  The historians were guided to annotate the ads using the same attributes of the English \textit{Runaways Slaves in Britain} dataset. See App. \ref{app:annotation guidelines} for annotation guidelines.
    
    Due to the expertise of the annotators and the annotation process being highly time-consuming, most ads were annotated by a single historian. Additionally, a random sample of 15 ads per language was annotated by a second annotator to calculate inter-annotator agreement (IAA) and assess the task's difficulty. The pairwise $F1$ agreement score \cite{tang-etal-2021-multi} for each language is calculated using the 15 dual-annotated ads, yielding high $F1$ scores of $91.5$, $83.2$ and $80.7$ for English, French and Dutch respectively. The higher agreement rate for English might be attributed to the cleaner source material in that language and possible differences in the complexity of the sources.  

    \textbf{In summary}, we now have annotated datasets in three languages -- the \textit{Runaway Slaves in Britain} in English randomly divided into train and validation splits, train sets in French and Dutch generated by translating the English train set, and manually annotated validation sets in French and Dutch.

\subsection{Models}
\label{sec:models}

\textbf{Ours} We experimented with several models trained with an extractive QA objective (see App. \ref{sec:training_details} for hyper-parameters) and evaluated them using the standard SQuAD-v2 $F1$ metric. We use standard RoBERTa-based monolingual models to be evaluated in monolingual settings, as it is a well-researched model known to achieve good performance on many downstream tasks and is available in English (RoBERTa), French \cite[CamemBERT;][]{martin2020camembert} and Dutch  \cite[RobBERT;][]{delobelle2020robbert}. We also test variations of these models, available in English, French and Dutch, that were successively fine-tuned on large extractive QA datasets. The English models were fine-tuned on SQuAD-v2, whereas the French models were fine-tuned on a collection of three datasets -- PIAF-v1.1 \cite{piaf}, FQuAD \cite{dhoffschmidt-etal-2020-fquad} and SQuAD-FR \cite{squad_fr}. The Dutch model was fine-tuned on SQuAD-NL, a machine-translated version of SQuAD-v2.\footnote{We translated it following the procedure described in \cite{squad_fr}.} In addition, we evaluate multilingual models of the XLM-RoBERTa \cite{xlm_roberta} family. We also test a variation of these models fine-tuned on SQuAD-v2. Finally, we investigate language models pre-trained on historical textual material, which are potentially better equipped to deal with historical ads. Specifically, we analyze the performance of MacBERTh \cite{jdmdh:9690}, a BERT-based model \cite{devlin-etal-2019-bert} that was pre-trained on historical textual material in English from 1450 to 1950. We also evaluate BERT models in English, French, and Dutch \cite{stefan_schweter_2020_4275044, stefan_schweter_dutch, stefan_schweter_english} that were trained specifically on historical newspapers from the 18th and the 19th centuries. Similarly, we also test variants of these models that were later fine-tuned on SQuAD.

\textbf{Baselines} We compare our models to two baselines suggested in prior work. \citet{de-toni-etal-2022-entities} used a T0++ model \cite{t0_2021multitask}, an encoder-decoder transformer with strong zero-shot capabilities, to perform NER tagging with historical texts in several languages. We adapt this to our task by converting the evaluation examples into prompts and feeding them into T0++ (See App. \ref{App:t0} for additional details). We also compare to OneIE \cite{lin-etal-2020-joint}, an English-only event extraction framework proposed by \citet{historical_event_extraction_2021}. 

Recall that \citet{liu-etal-2020-event} also constructed event extraction as a QA task. However, their model cannot be directly compared to ours -- \citeauthor{liu-etal-2020-event} supports only single sentences, while we process entire paragraphs; and adapting their model to new events which do not appear in their training dataset (as in our case) would require extensive effort, specifically for the multilingual settings. 
We thus leave such an investigation for future work.

\subsection{Experimental Setup}
\label{sec:setup}

The main goal of this paper is to determine the most successful approach for event extraction from historical texts with varying resources (e.g. the number of annotated examples or the existence of datasets in various languages). We therefore evaluate the models described in §\ref{sec:models} with the following settings.

    \textbf{Zero-shot inference} 
    This simulates the prevalent case for historical NLP where no in-domain data is available for training.
    
    \textbf{Few-shot training} Another frequent setup in the historical domain is where experts labeled a small number of training examples. Therefore, we train the models on our annotated monolingual datasets of various sizes (from a few examples to the entire dataset) and test their performance on evaluation sets in the same language. 
    
    \textbf{Semi-supervised training} Sometimes, in addition to a few labeled examples, a larger unlabeled dataset is available. We thus also evaluate our monolingual models in semi-supervised settings, where we either: 1) further pre-train the models with a masked language modeling objective (MLM) using the unannotated dataset, then fine-tune them on our annotated dataset; 2) simultaneously train the models with an MLM objective using the unannotated dataset and on the standard QA objective using the annotated dataset; or 3) use an iterative tri-training \cite{tri_training} setup to utilize the larger unannotated dataset. In tri-training, three models are trained on a labeled dataset and are used to predict the labels of unlabeled examples. All the samples for which at least two models agree on are added to the labeled set. Finally, a new model is trained on the resulting larger labeled dataset.  
    
    \textbf{Cross-lingual training} Finally,
    we test two cross-lingual training variations. In the simple setting, we train a multilingual model on the labeled English dataset, evaluating it on French or Dutch. In the MLM settings, we also train the model with an MLM objective using the unlabeled target data.

\section{Results and Analysis}
\label{sec:results}


\subsection{Zero-Shot Inference} 
\label{sec:zero_shot}


\begin{table}[t]
\fontsize{10}{10}\selectfont
 \begin{tabular}{p{3.5cm}p{2cm}c}
    \toprule
    Model & Fine-tune data  & $F1$ \\ \midrule
    \multicolumn{3}{l}{\textbf{en}}  \\ \midrule
    OneIE & N\textbackslash A & 51.90 \\
    T0++ & N\textbackslash A & 33.69 \\ 
    RoBERTa-base & SQuAD-v2 & $54.35$ \\  %
    RoBERTa-large & SQuAD-v2 & $\textbf{56.42}$\\ %
    XLM-RoBERTa-base & SQuAD-v2 & $41.84$ \\  %
    XLM-RoBERTa-large & SQuAD-v2 & $55.10$ \\ \midrule %
    \multicolumn{3}{l}{\textbf{fr}}  \\ \midrule
    T0++ & N\textbackslash A & 32.26 \\
    CamemBERT-base & PIAF-v1.1   FQuAD-v1   SQuAD-FR & $30.65$ \\  %
    XLM-RoBERTa-base & SQuAD-v2 & $36.51$ \\ %
    XLM-RoBERTa-large & SQuAD-v2 & $\textbf{44.52}$ \\ \midrule %
    \multicolumn{3}{l}{\textbf{nl}}  \\ \midrule
    T0++ & N\textbackslash A & 29.28 \\ 
    RobBERT-base & SQuAD-NL & $37.21$ \\
    XLM-RoBERTa-base & SQuAD-v2 & $37.56$ \\ %
    XLM-RoBERTa-large & SQuAD-v2 & $\textbf{40.42}$ \\
    \bottomrule %

 \end{tabular}
 \caption{Zero-shot performance of different models.}
 \label{tab:zero_shot}
\end{table}

Tab. \ref{tab:zero_shot} demonstrates the benefit of framing event extraction as extractive QA. Indeed, almost all the QA models outperform the T0++ baseline by a large margin. Most English models also have significant gains over OneIE.
As can also be observed from the table, the overall performance is much better for English compared to Dutch and French. This performance gap can likely be attributed to differences in the sources from which the datasets were curated. The higher IAA for the English dataset (§\ref{sec:datasets}) further supports this hypothesis. In addition, since English is the most high-resource language \cite{wu-dredze-2020-languages}, models trained on it are expected to perform best. This difference in availability of resources might also explain why the multilingual models perform better than the monolingual models on French and Dutch, while the monolingual models outperform the multilingual ones for English \cite{rust-etal-2021-good}.
Unsurprisingly, it can also be seen that the larger LMs achieve significantly higher $F1$ scores compared to the smaller models.

\subsection{Few-Shot Training}

      
      


\begin{figure*}[t]
    \centering
    \begin{subfigure}{0.49\textwidth}
        \centering
        \includegraphics[width=\textwidth, center, trim={0.2cm 0.45cm 0.35cm 0.25cm},clip]{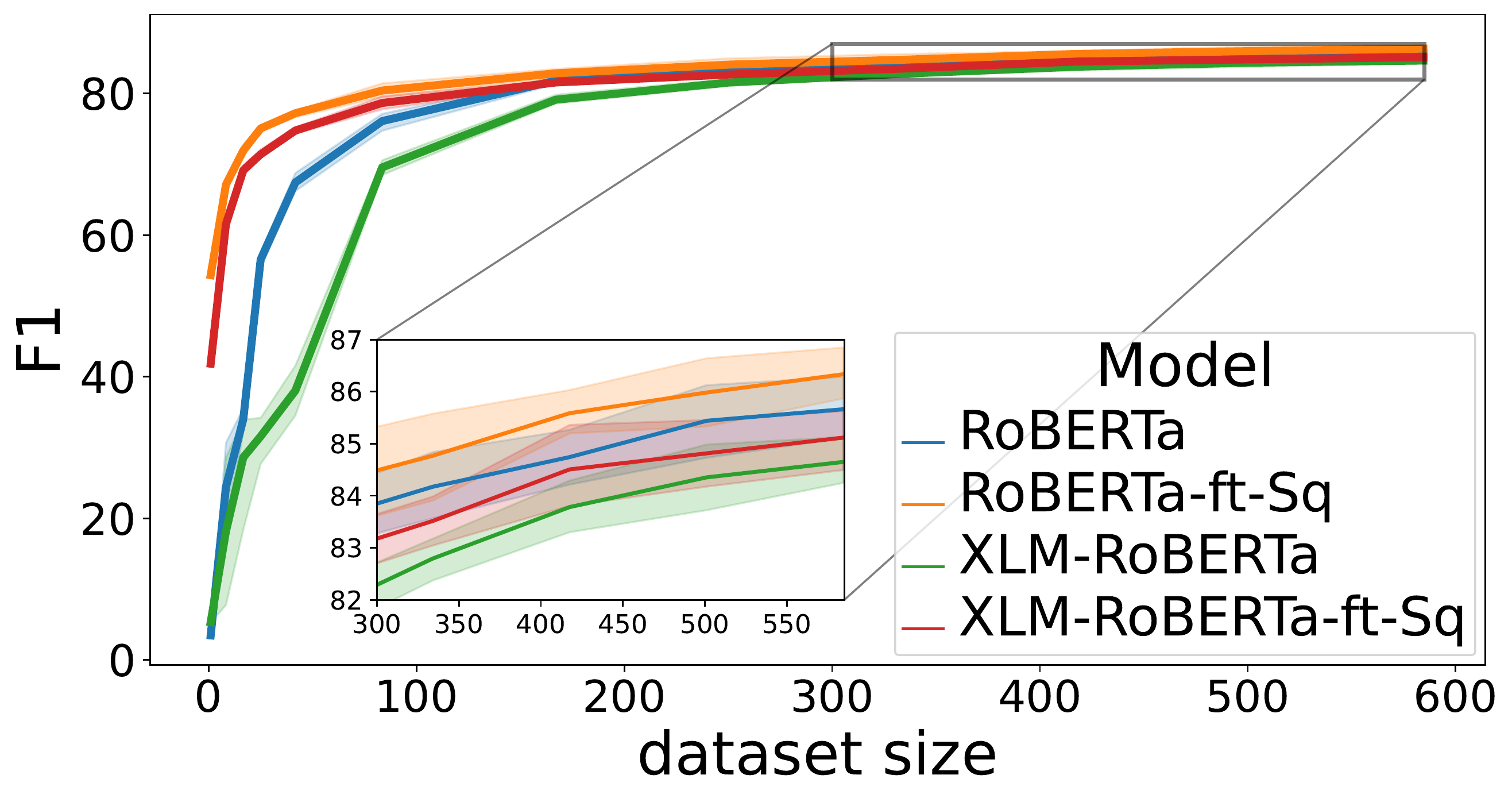}
        \caption{English modern}
        \label{fig:english_few_shot}
      \end{subfigure}%
       \begin{subfigure}{0.49\textwidth}
        \centering
        \includegraphics[width=\textwidth, center, trim={0.2cm 0.45cm 0.35cm 0.25cm},clip]{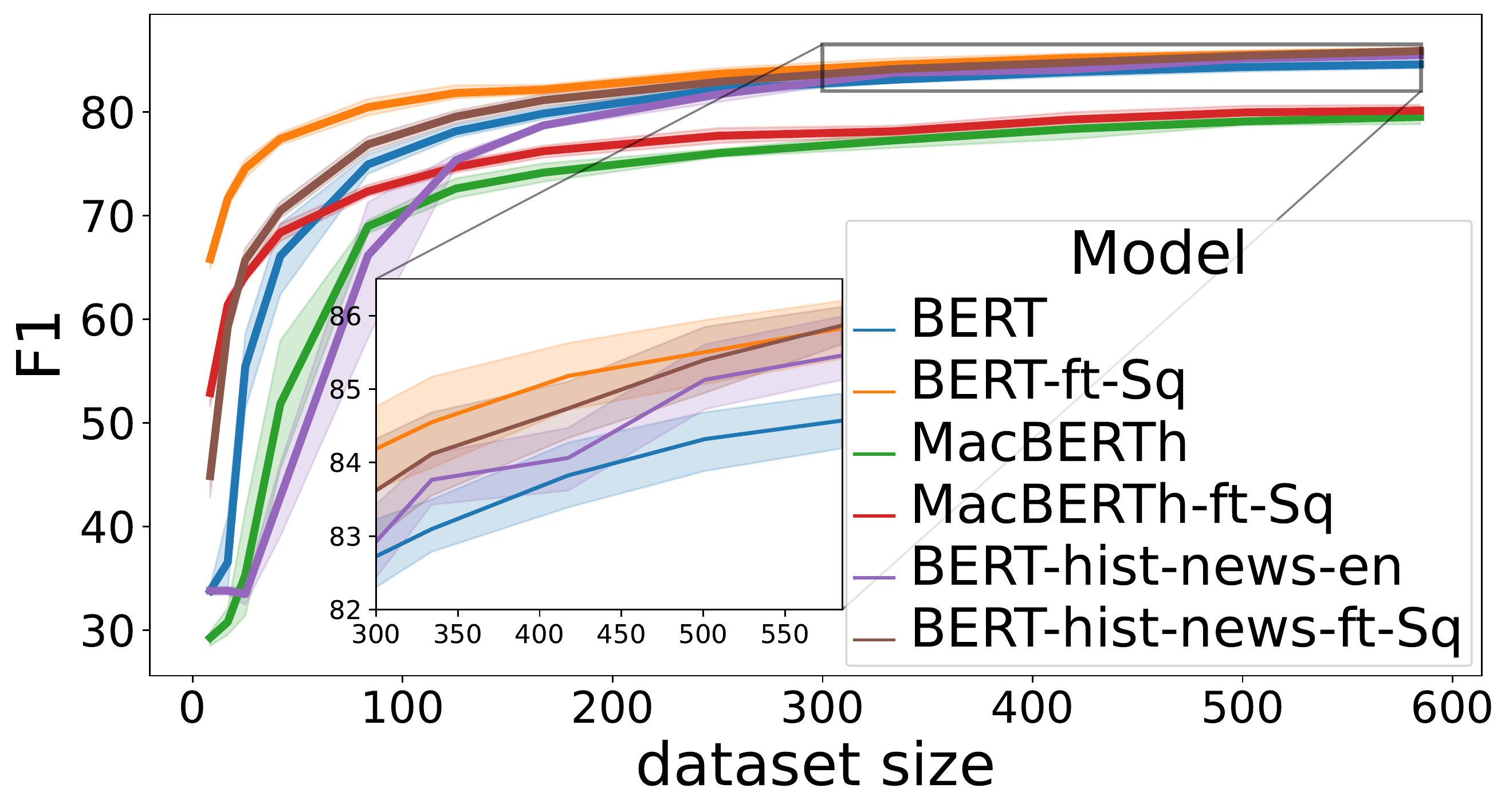}
        \caption{English historical}
        \label{fig:english_historical}
      \end{subfigure}%
      
      \begin{subfigure}{0.49\textwidth}
        \centering
        \includegraphics[width=\textwidth, center, trim={0.2cm 0.45cm 0.35cm 0.25cm},clip]{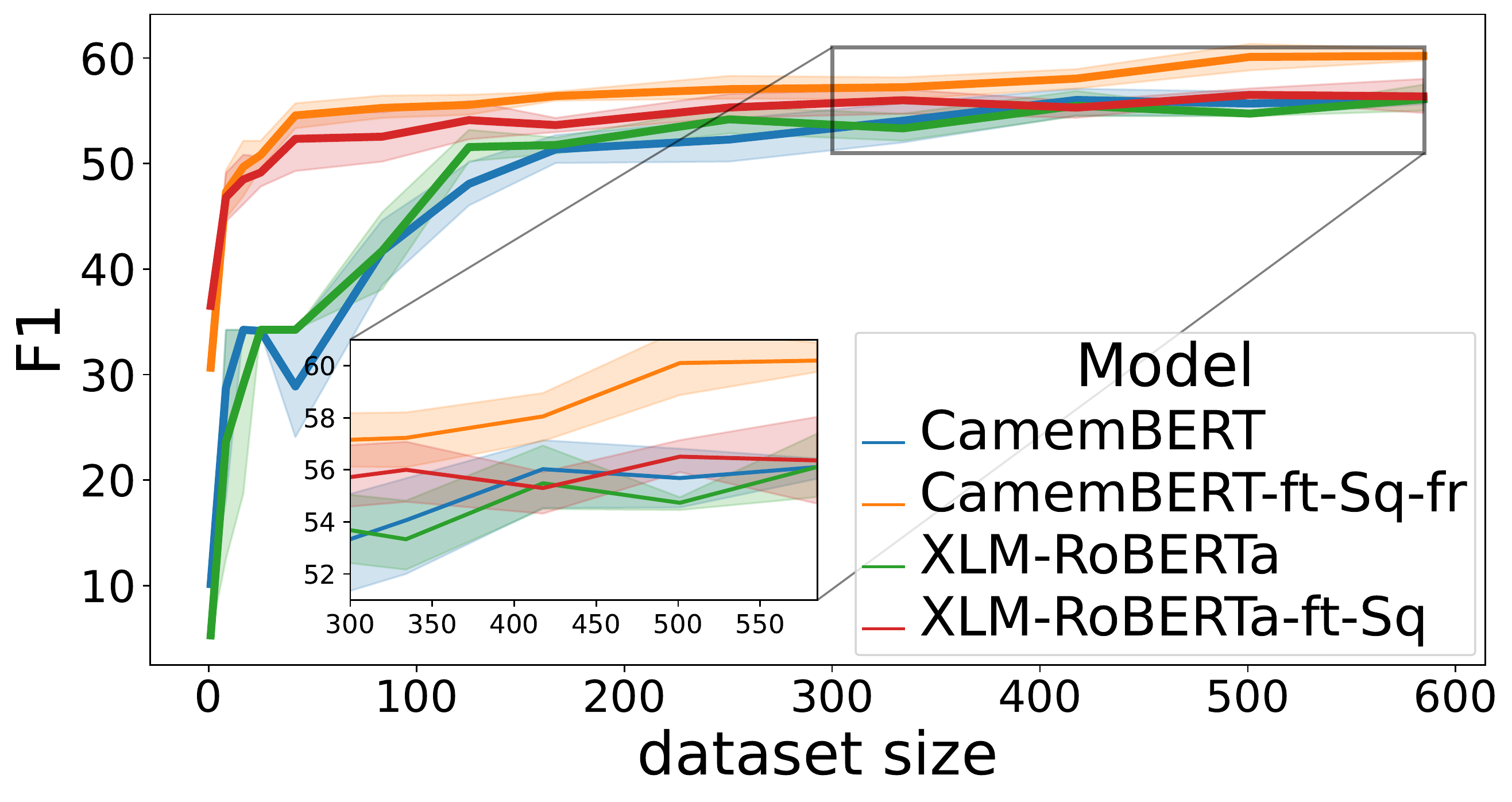}
        \caption{French modern}
        \label{fig:french_few_shot}
      \end{subfigure}%
      \begin{subfigure}{0.49\textwidth}
        \centering
        \includegraphics[width=\textwidth, center, trim={0.2cm 0.45cm 0.35cm 0.25cm},clip]{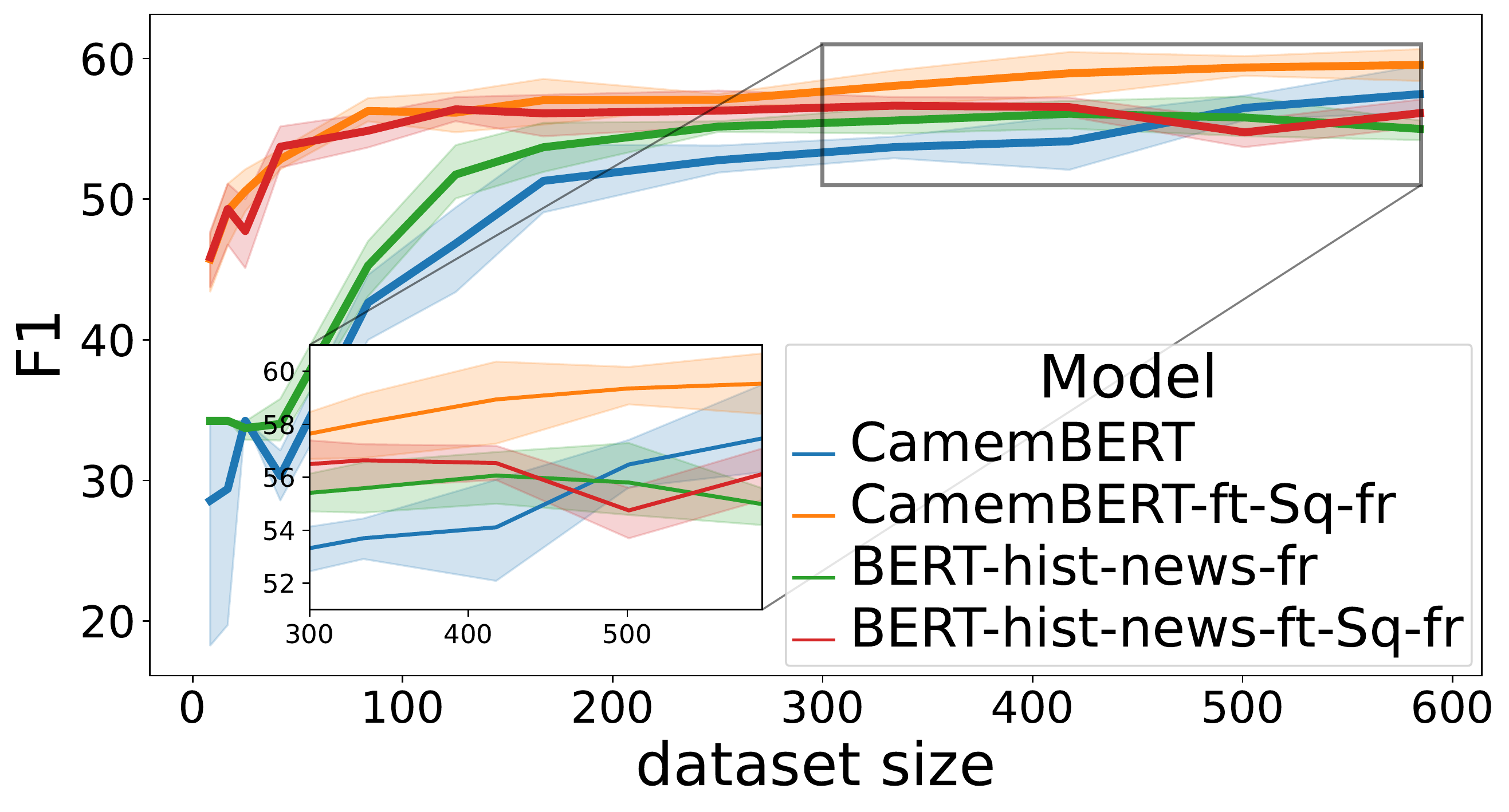}
        \caption{French historical}
        \label{fig:french_historical}
      \end{subfigure}%
      
      \begin{subfigure}{0.49\textwidth}
        \centering
        \includegraphics[width=\textwidth, center, trim={0.2cm 0.45cm 0.35cm 0.25cm},clip]{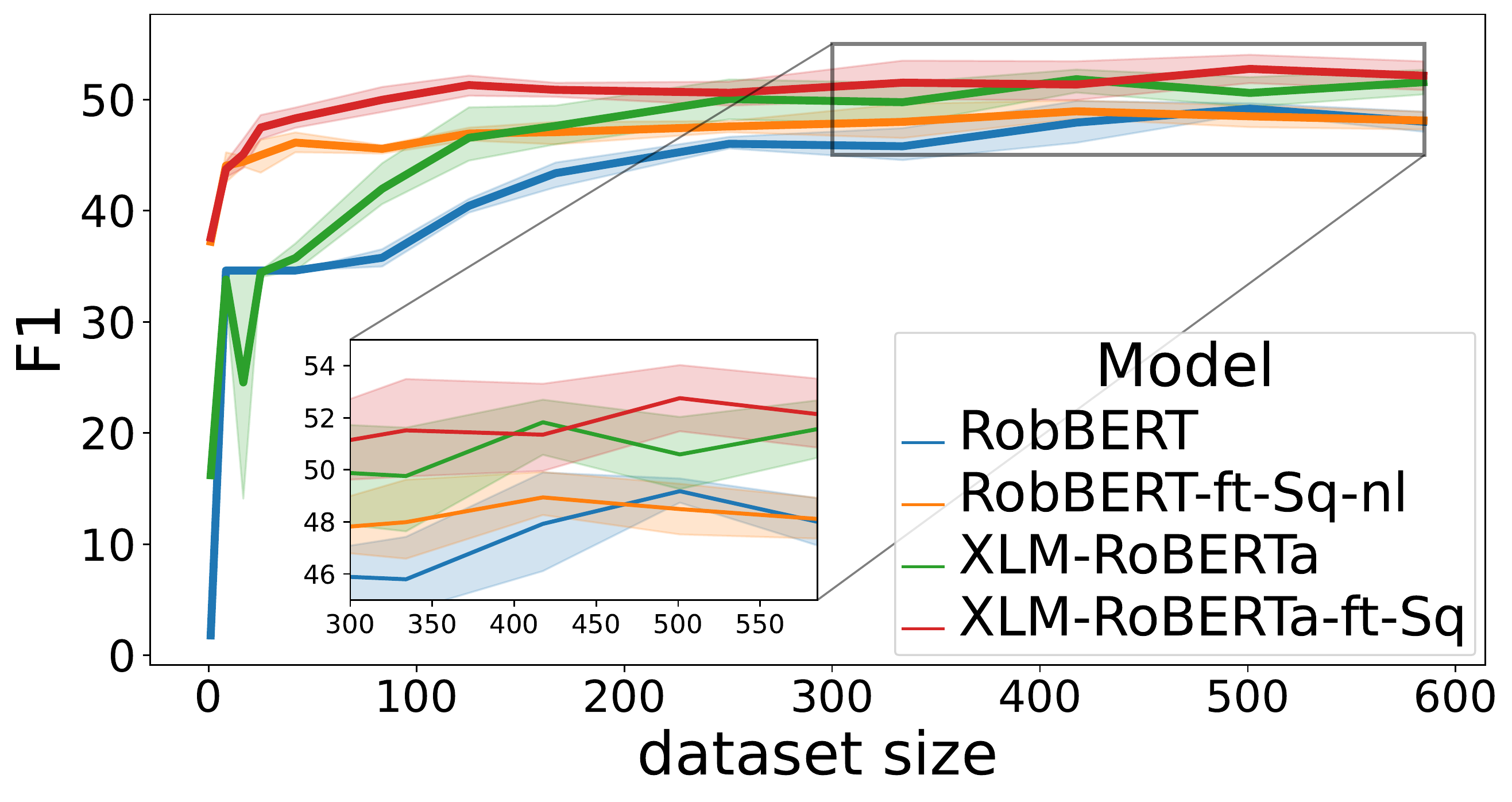}
        \caption{Dutch modern}
        \label{fig:dutch_few_shot}
      \end{subfigure}
    \centering
      \begin{subfigure}{0.49\textwidth}
        \centering
        \includegraphics[width=\textwidth, center, trim={0.2cm 0.45cm 0.35cm 0.25cm},clip]{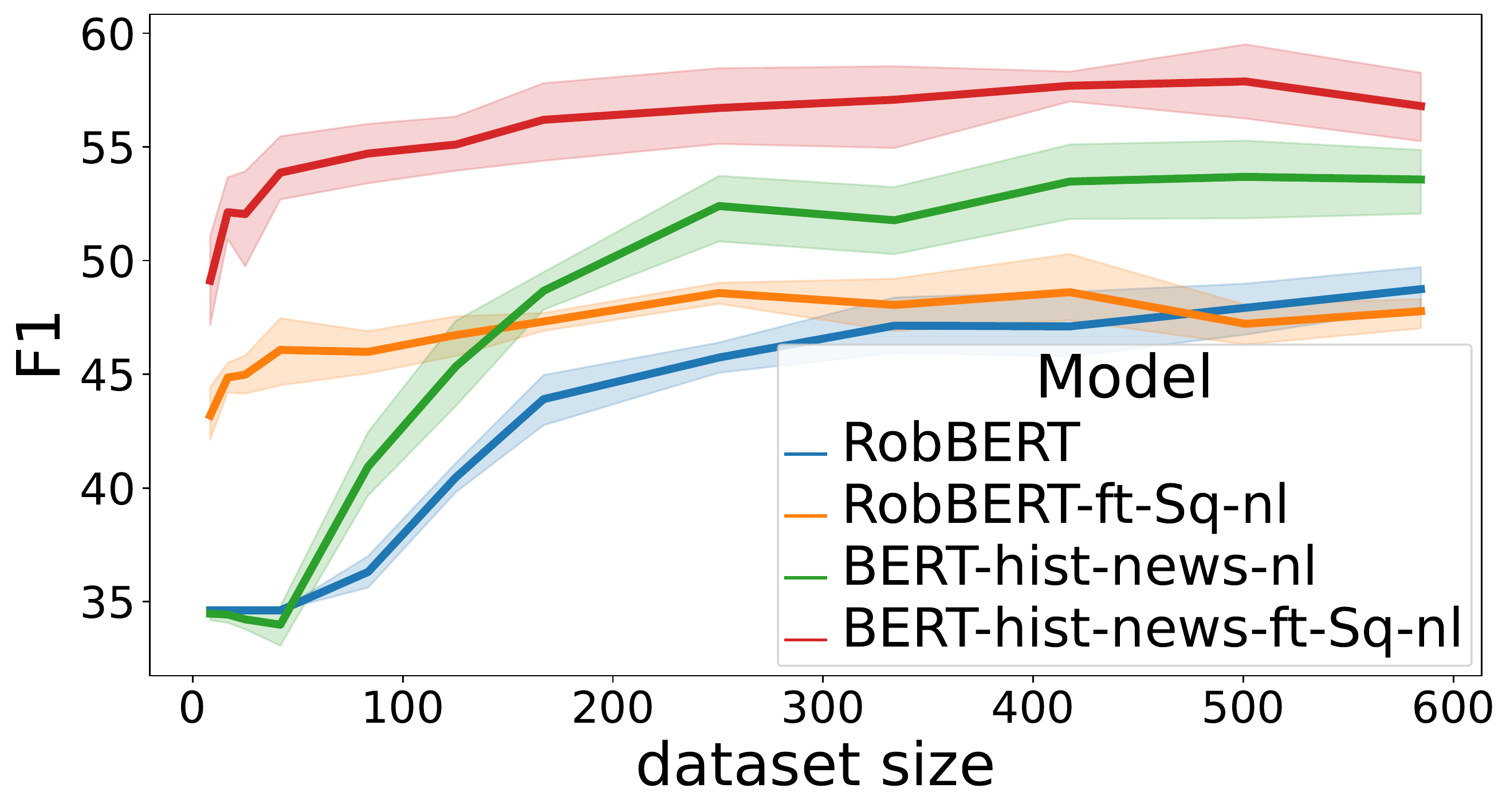}
        \caption{Dutch historical}
        \label{fig:dutch_historical}
      \end{subfigure}

    \caption{Performance of the models in a few-shot setting for the three languages, historical and modern models. All models were trained using their ``base" version. ``ft-Sq'' signifies that the model was fine-tuned on SQuAD or one of its equivalents in French (fr) or Dutch (nl).}%
    \label{fig:historical}%
\end{figure*}

\begin{table*}[t]
    \centering
    \fontsize{10}{10}\selectfont
    \begin{tabular}{cllcccc}
    \toprule
        \multirow{2}{*}{Language} & \multirow{2}{*}{Model} & \multirow{2}{*}{Setting} & \multicolumn{4}{c}{Dataset size} \\  \cmidrule{4-7}
         ~ & ~ & ~ & 8 & 16 & 25 & 585 \\ \midrule

        \multirow{4}{*}{en} & \multirow{4}{*}{RoBERTa-base-ft-SQuAD} & None & 67.13 & 77.2 & 80.41 & 86.33 \\  
        ~ & ~ & Further pre-trained & 57.18 & 76.52 & 79.93 & 85.91 \\ 
        ~ & ~ & MLM semi-supervised & 68.28 & 78.17 & 80.8 & 86.17 \\
        ~ & ~ & Tri-training & \underline{\textbf{70.97}} &\underline{\textbf{79.48}} & \underline{\textbf{82.42}} & \underline{\textbf{87.04}} \\ \midrule 
        \multirow{7}{*}{fr} & \multirow{4}{*}{CamemBERT-base-ft-SQuAD} & None & \underline{\textbf{47.3}} & \underline{\textbf{54.55}} & 55.26 & 60.19 \\
        ~ & ~ & Further pre-trained & 34.04 & 49.48 & 54.04 & 61.01 \\ 
        ~ & ~ & MLM semi-supervised & 46.79 & 48.2 & 47.11 & 49.64 \\ 
        ~ & ~ & Tri-training & 46.76 & 53.87 & \underline{\textbf{55.98}} & \underline{\textbf{61.58}} \\ \cmidrule{2-7}
        ~ & \multirow{3}{*}{XLM-RoBERTa-base-ft-SQuAD} & None & 46.8 & 48.48 & 49.14 & \underline{56.36} \\
        ~ & ~ & Simple cross-lingual & 46.08 & \underline{51.01} & \underline{51.45} & 56.28 \\ 
        ~ & ~ & MLM cross-lingual & \underline{47.0} & 48.36 & 48.34 & 53.98 \\  \midrule
        \multirow{6}{*}{nl} & \multirow{3}{*}{RobBERT-base-ft-SQuAD} & None & \underline{44.04} & 46.12 & 45.56 & 48.11 \\  
        ~ & ~ & Further pre-trained & 34.61 & \underline{\textbf{46.16}} & \underline{\textbf{48.15}} & \underline{49.84} \\ 
        ~ & ~ & MLM semi-supervised & 31.6 & 41.62 & 40.22 & 43.82 \\ \cmidrule{2-7}
        ~ & \multirow{3}{*}{XLM-RoBERTa-base-ft-SQuAD} & None & 43.73 & 45.08 & \underline{47.47} & \underline{\textbf{52.14}} \\ 
         ~ & ~ & Simple cross-lingual & 43.32 & 44.84 & 44.79 & 46.63 \\
         ~ & ~ & MLM cross-lingual & \underline{\textbf{45.94}} & \underline{45.34} & 47.1 & 48.5 \\ 
        
        \bottomrule
    \end{tabular}
    \caption{$F1$ score of the models in semi-supervised and cross-lingual settings. ``None'' means the model was trained in a standard supervised fashion. For ``further pre-trained'' we first further train the model on an MLM objective, then train it on our annotated dataset. For ``MLM semi-supervised'' we train the models on MLM and QA objectives simultaneously, and in ``tri-training'' we train the models using the tri-training algorithm. This line is missing from the Dutch models as the unlabeled Dutch dataset contains entire newspaper issues and not individual ads. `Simple cross-lingual'' is standard cross-lingual training and ``MLM cross-lingual'' marks that the model was trained using an MLM-objective in addition to the standard QA loss. Bold marks the best method for a language, while an underline marks the best method for a specific training setting (semi-supervised or cross-lingual).
See Tab. \ref{tab:semi_supervised_app} and \ref{tab:cross_lingual_app} in App. \ref{app:additional results} for evaluation of other models.}
    \label{tab:semi_supervised}%
\end{table*}

Next, we analyze the results of fine-tuning the models in a fully supervised setting in a single language. Fig. \ref{fig:english_few_shot} shows the performance of four models on the English evaluation set after being fine-tuned on English training sets of various sizes. All models achieve impressive $F1$ scores even when trained on a small fraction of the training set, further demonstrating the benefit of formulating the task as an extractive QA problem.

Interestingly, the two models intermediately trained on SQuAD perform better than the base models. This trend holds for all dataset sizes but is particularly pronounced in the low-data regime, demonstrating that the SQuAD-based models can generalize with much fewer examples. Comparing Fig. \ref{fig:english_few_shot} with Tab. \ref{tab:zero_shot} further underpins this finding. In addition, we again see that the multilingual models achieve lower $F1$ scores than their monolingual counterparts. Moreover, and unsurprisingly, our results also suggest that the large models perform better than their base versions (Fig. \ref{fig:model_sizes} in App. \ref{app:additional results}).

Fig. \ref{fig:french_few_shot}, \ref{fig:dutch_few_shot} repeat some of the trends mentioned above and in §\ref{sec:zero_shot}. Again, the models achieve considerably lower $F1$ scores in French and Dutch than in English. While our evaluation of the translation demonstrated the relatively high quality of the process, This gap can still be attributed to noise in the translation process of the train datasets from English to Dutch and French and its bias towards modern language.
In addition, for both French and Dutch, the SQuAD-fine-tuned models reach higher $F1$ scores for most (but not all) dataset sizes. Fig. \ref{fig:dutch_few_shot} demonstrates, similar to  Tab. \ref{tab:zero_shot}, that multilingual models perform better than the monolingual models for Dutch. Surprisingly, this result cannot be observed in Fig. \ref{fig:french_few_shot}: A monolingual French model outperforms the two multilingual models by a large margin. Finally, we again see (Fig. \ref{fig:model_sizes}) that larger language models achieve better results than their smaller versions.

We now investigate language models pre-trained on historical texts and find surprising results (Fig. \ref{fig:historical}). MacBERTh performs worse than BERT,\footnote{For the purpose of fairness, we use BERT rather than RoBERTa for comparison with MacBERTh and BERT-hist-news-en, which are BERT-based models.} despite being trained on historical English texts. However, BERT-hist-news-en, trained on historical newspapers, performs better on some data regimes. We further analyze this in §\ref{sec:analysis}.

The analysis of the French models reveals a slightly different picture (Fig. \ref{fig:french_historical}). However, directly comparing CamemBERT and BERT-hist-news-fr is not possible, as the former is based on RoBERTa while the latter is based on BERT. The results for the Dutch models, presented in Fig. \ref{fig:dutch_historical}, are particularly intriguing. BERT-hist-news-nl performs significantly better than RobBERT, to the extent that the difference cannot be solely attributed to the differing architectures of the two models.\footnote{RobBERT is based on RoBERTa and BERT-hist-news-nl is based on BERT.} As XLM-RoBERTa also outperforms RobBERT, it seems that this model may not be well-suited for this specific domain. These findings will be further explored in §\ref{sec:analysis}.

\subsection{Semi-Supervised Training}

Tab. \ref{tab:semi_supervised} reveals an interesting result: for English, using the larger unannotated dataset improved the performance of the models for all data sizes. Moreover, tri-training is most effective for English. 
The picture is less clear, however, for French and Dutch. While using the unannotated data has a positive impact on models trained on the entire dataset, the gains are smaller and tend to be unstable. We leave an in-depth exploration of this for future work. 

\subsection{Cross-lingual Training}

As mentioned in §\ref{sec:setup}, we compare two different cross-lingual settings: supervised-only
, where we train a cross-lingual model on the English \textit{Runaway Slaves in Britain} dataset while evaluating it on French or Dutch; and MLM settings, where we also train the model with an MLM-objective using an unlabeled dataset of the target language. Tab. \ref{tab:semi_supervised} contains the results of this evaluation. Interestingly, it seems that cross-lingual training is more effective when the number of available annotated examples is small. When the entire dataset is being used, however, monolingual training using a translated dataset achieved better performance. Tab. \ref{tab:semi_supervised} also demonstrates that the MLM settings are preferable over the simple settings in most (but not all) cases.   

\subsection{Error Analysis}
\label{sec:analysis}
\begin{figure*}

\centering
     \includegraphics[width=\textwidth, center]{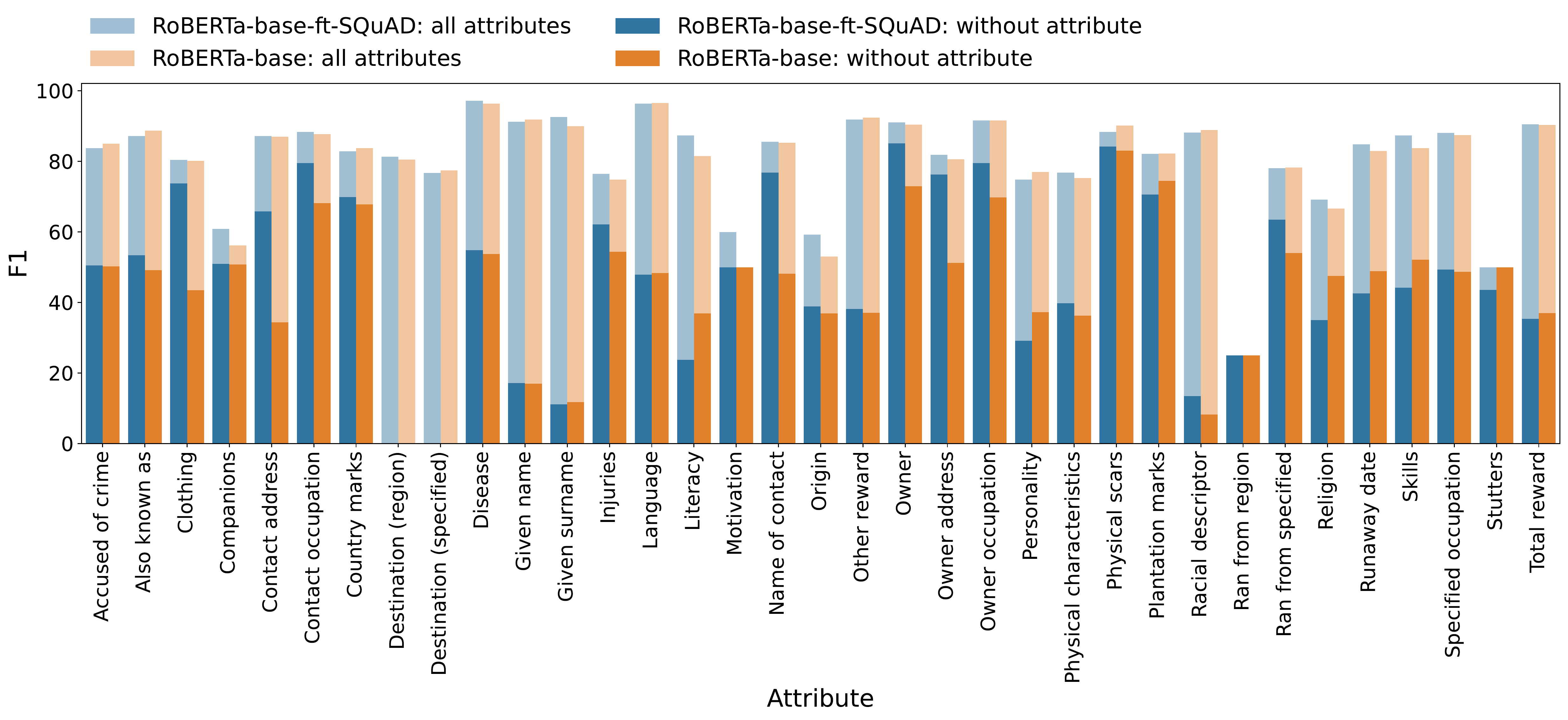}
     \caption{The generalization capabilities of RoBERTa in a fully-supervised setting. The columns in pale color describe the performance of the models on the attribute with standard training, whereas the columns in darker color describe the performance on the attribute of a model that was not trained on the attribute (generalization).}
     \label{fig:gen_high}
\end{figure*}

First, we investigate common errors that our most successful models (RoBERTa) make. Fig. \ref{fig:ad_length} in App. \ref{app:additional results} demonstrates that the model struggles with long ads. 
Perhaps using models that were trained on longer sequences could help with this going forward. A per-attribute analysis, the result of which can be seen in Fig. \ref{fig:gen_high} (pale-colored columns), unsurprisingly suggests that the model finds rare attributes harder to predict (e.g. ``ran from region'', and compare Fig. \ref{fig:gen_high} to Tab. \ref{tab:attributes}). 

Next, we move on to evaluating the generalization capabilities of the models. A per-attribute analysis (Fig. \ref{fig:gen_high}, dark-colored columns) reveals that training RoBERTa on SQuAD improved the overall ability of the model to generalize to unseen attributes, probably by utilizing the much broader types of questions that exist in the dataset. However, we also see that the models particularly struggle to generalize to some of them. After closer examination, it seems like these ``hard'' attributes are either: 1) very rare (``Destination (region)''); 2) non-specific, with possibly more than one span in the ad with the correct type of the answer (``Given name''); or 3) related to topics that are probably not being represented in SQuAD (``Racial descriptor''). We speculate that a more well-tuned conversion of the attributes to natural questions could mitigate some of these issues. 

Finally, we compare historical LMs to modern models to understand why MacBERTh underperforms on the \textit{Runaways Slaves in Britain} dataset while BERT-hist-news-en/nl do not. We hypothesize that MacBERTh, trained on a wide range of texts from over 500 years, cannot adapt well to ads written in a language more similar to modern English. Additionally, MacBERTh's training dataset is disproportionately skewed towards texts from 1600-1690 and 1830-1950, while texts from 1700-1850 (the period corresponding to our dataset) are scarce. In contrast, BERT-hist-news-en/nl were trained on datasets containing mostly 19th-century newspapers, a domain and period closer to our.

To validate this, we calculate the perplexity of our dataset w.r.t. the models (technical details in App. \ref{App:perplexity}). Indeed, the perplexity of our English newspaper ads dataset w.r.t. MacBERTh is higher ($16.47$) than the perplexity w.r.t. BERT ($15.32$) and BERT-hist-news-en ($5.65$). A similar picture emerges for Dutch:  the perplexity of our Dutch test dataset of newspaper ads w.r.t RobBERT was significantly higher ($49.53$) than the perplexity w.r.t. BERT-hist-news-nl ($5.12$).   


\section{Conclusions}
\label{sec:conclusions}
In this work, we address the unique challenges of event extraction from historical texts in different languages. We start by developing a new multilingual dataset in English, French, and Dutch of events, consisting of newspaper adverts reporting on enslaved people escaping their enslavers. We then demonstrate the benefits of framing the problem as an extractive QA task. We show that even with scarcely annotated data, this formulation can achieve surprisingly good results by leveraging existing datasets and models for modern languages. Finally, we show that cross-lingual low-resource learning for historical languages is highly challenging, and machine translation of the historical datasets to the considered target languages is, in practice, often the best-performing solution.


\section*{Limitations}
\label{sec:limitations}
We see four main limitations regarding our work. First, we have evaluated our models on a dataset containing events of one type only. It remains to be seen how applicable our formulation and methods are to other historical datasets and event types. 
Second, given the nature of the historical question our dataset targets, it contains documents only from one language family. Extending our methodology to languages from other language families might pose further challenges in terms of multilinguality. 
Third, our method relies heavily on automatic translation tools, which are biased toward translating historical texts into modern language. This can negatively affect the performance of our models.  
Lastly, in real-life cases, machine readable historical texts are often extremely noisy, suffering from high level of OCR errors and other text extraction mistakes. Conversely, we have tested our methods on relatively clean datasets, with the unannotated Dutch material as the only exception. We leave a more thorough study on how well our proposed methods are suitable for noisy text to future work.


\section*{Ethical Considerations}
\label{sec:ethics}
Studying texts about the history of slavery poses ethical issues to historians and computer scientists alike since people of color still suffer consequences of this history in the present, not least because of lingering racist language \cite{alim_raciolinguistics_2016, alim_oxford_2020}.

As researchers, we know that an important ethical task is to develop sound NLP tools that can aid in the examination of historical texts containing racist language, while endeavoring at all costs not to reproduce or perpetuate such racist language through the very tools we develop. 

The enslaved people described in the newspapers adverts used in this study were alive centuries ago, so any immediate issues related to their privacy and personal data protection do not apply. Nonetheless, the newspaper adverts studied here were posted by the oppressors of the people who tried to liberate themselves, and contain many examples of highly racist and demeaning language. 



\bibliography{anthology,custom}
\bibliographystyle{acl_natbib}

\clearpage
\appendix

\section*{Appendix}
\label{sec:appendix}
\section{Reproducibility}

\subsection{Calculating Perplexity}
\label{App:perplexity}

To calculate the (pseudo)-perplexity of a sentence $S = w_1 w_2 w_3 ... w_n$ w.r.t. a masked language model, we used the following formula

\begin{equation}
    \begin{split}
        PP_{(S)} &= \left( \prod_{i=1}^{n}P(w_{i} | S_{-i})\right)^{-1/n} \\ 
        &= \left( \prod_{i=1}^{n}P_{\textrm{MLM}}(w_{i} | S_{-i})\right)^{-1/n}
    \end{split}
\end{equation}

where $S_{-i}$ is the sentence $S$ masked at token $i$. To calculate the perplexity of an entire corpus $C = {S^1, S^2,..., S^m}$ w.r.t. a masked language model we notice that $P(w^j_i|C_{-(j,i)}) = P(w^j_i|S^j_{-i})$, where $C_{-(j,i)}$ is the corpus $C$ with sentence $j$ masked at location $i$.

Therefore,

\begin{equation}
    PP_{(C)} = \left(\prod_{j=1}^{m} \prod_{i=1}^{|S^j|}P_{\textrm{MLM}}(w^j_{i} | S^j_{-i})\right)^{-1/k}
\end{equation}

where $k$ is the total number of tokens in the corpus, i.e. $k = \sum_{j=1}^m |S^j|$ .

Notice, that in the log space this formula becomes equivalent to the average of the negative log likelihoods:

\begin{equation}
    \log{(PP_{(C)})} = \frac{1}{k}\left(\sum_{j=1}^{m} \sum_{i=1}^{|S^j|}\textrm{NLL}_{\textrm{MLM}}(w^j_{i} | S^j_{-i})\right)
\end{equation}

where $\textrm{NLL}_{\textrm{MLM}}$ is the negative log likelihood, which in many cases equal to passing the output of the language model to a standard cross entropy loss.

\subsection{Translation of the Annotated Dataset}
\label{App:translation}

\subsubsection{Translation Process}

Each sample in the annotated Runaways dataset follows the SQuAD-v2 scheme, and contains a context $c$ (the ad's text), a question $q$ (one of the attributes) and an answer $a$ such that $a$ appears in $c$ ($a$ might also be the empty string). We used the publicly available Google Translate API\footnote{using the deep-translator package, \url{https://deep-translator.readthedocs.io/en/latest/}} to translate the samples into the target languages. We also considered using Facebook's NLLB model \cite{costa2022no},\footnote{we used the $3.3$b parameters variant \url{https://huggingface.co/facebook/nllb-200-3.3B}, as it was the biggest model available we could load on our machine} but it performed noticeably worse. See below for more details regarding evaluating the quality of the translation.

Unfortunately, simply translating $(c, q, a)$ from English to the target language is not enough. In some cases, translation of the context and the answer are not always aligned. That is, translating $c$ to $c^t$ and $a$ to $a^t$ results in a pair for which $a^t$ does not appear verbatim in $c^t$. In those cases we try to find a span of text $\hat{a}^t$ in $c^t$ such that $\hat{a}^t$ is similar to $a^t$ (and therefore, hopefully the correct answer to the question $q$).

To achieve this, we use fuzzy string matching\footnote{using \url{https://pypi.org/project/fuzzywuzzy/}} to find $\hat{a}^t$. Specifically, we did the following. First, we calculated $k = \max(|a^t|, |a|)$, and extracted all the k-grams from $c^t$. Then, we used fuzzy string search to find the k-gram that is most similar to $a^t$, with a score of at least 0.5. We then assign $k = k + 1$ and repeat the process five times, finally returning the match with the highest score. If no match was found, we assign $a^t = a$ (this is useful in cases where the answer is a name, a date etc.) and repeat the above-mentioned algorithm. If again no match is found the matching has failed and we discard the sample.   

Finally, we opted to manually translate $q$ as the number of different questions in our dataset is relatively low.

\subsubsection{Evaluation of the Translation}

\begin{table}[t]
\fontsize{10}{10}\selectfont
 \begin{tabular}{llc}
    \toprule
    Language & Translation tool  & COMET score \\ \midrule
    \multirow{2}{*}{French} & Google Translate & \textbf{0.014} \\
                            & NLLB & 0.01 \\ \midrule
    \multirow{2}{*}{Dutch} & Google Translate & \textbf{0.017} \\
                            & NLLB & 0.01 \\
    \bottomrule %

 \end{tabular}
 \caption{Evaluation of the translation quality using COMET (higher is better).}
 \label{tab:translation_evaluation_comet}
\end{table}

\begin{table}[t]
\fontsize{10}{10}\selectfont
 \begin{tabular}{llcc}
    \toprule
    Language & Translation tool  & Accuracy & Fluency \\ \midrule
    \multirow{2}{*}{French} & Google Translate & \textbf{4.5} & \textbf{3.4}\\
                            & NLLB & 3.7 & \textbf{3.4} \\ \midrule
    \multirow{2}{*}{Dutch} & Google Translate & \textbf{4.8} & \textbf{4.2}\\
                            & NLLB & 3.5 & 3.3 \\
    \bottomrule %

 \end{tabular}
 \caption{Evaluation of the translation quality using human raters (higher is better).}
 \label{tab:translation_evaluation_human}
\end{table}

We evaluated several translation tools. Based on preliminary evaluation, we determined that Google Translate and Facebook's NLLB model were the most promising options, as other methods either did not meet the minimum desired quality or were difficult to run on large datasets. We evaluated the two translation schemes using automatic tools and human raters. Both metrics demonstrated the superiority of Google Translate over NLLB in terms of accuracy and fluency, as shown below.

\textbf{Automatic method} We used COMET, a state-of-the-art reference-free automatic translation evaluation tool \cite{rei-etal-2021-references}, and used it to evaluate the quality of translating the original English ads to French and Dutch. Tab. \ref{tab:translation_evaluation_comet} contains the result of running the model, demonstrating the higher quality of the translations produced by Google Translate compared to NLLB.

\textbf{Human evaluation} We asked native speakers to rate 20 translations of ads on a scale of 1-5 for accuracy and fluency. They were instructed to give a translation a fluency score of 5 if it is as fluent as the original English text, and 1 if it was barely readable. Similarly, they were instructed to give an accuracy score of 5 if all the ad's attributes describing the self-liberation event were translated correctly and 1 if almost none of them were. Tab. \ref{tab:translation_evaluation_human} demonstrate not only that Google Translate is the better translation tool, but also that the accuracy and fluency of the tool are objectively good.


\subsection{Zero-Shot Inference with T0++}
\label{App:t0}

T0++ is a prompt-based encoder-decoder LM developed as part of the BigScience project \cite{t0_2021multitask}. One of the tasks that T0++ was trained on is extractive QA. To train the model on an extractive QA task, the designers of T0++ converted an extractive QA dataset, such as SQuAD into a prompt format. Each example with question $q$, context $c$ and answer $a$ in the dataset was placed into one of several possible templates, such as ``\textit{Given the following passage: \{$c$\}, answer the following question. Note that the answer is present within the text. Question: \{$q$\}}''. T0++ was trained to generate $a$ given the template as a prompt. 

To perform inference with T0++ with our datasets we followed \citet{de-toni-etal-2022-entities} and the original training routine of T0++. We converted the dataset to prompts using one of the templates that were used to train the model on extractive QA, and tried to map T0++'s prediction into the original context. As \citet{de-toni-etal-2022-entities} we tried two mapping methods -- an exact matching, where we consider T0++'s prediction valid only if the prediction appears verbatim in the context; and a fuzzy matching method, where some variation is allowed. If no match is found we discard the prediction and assume that the answer to the question does not exist in the context. In Tab. \ref{tab:zero_shot} we report the result of the ``exact match'' method, which performed better in practice.

\subsection{Training Details}
\label{sec:training_details}
We specify here the hyper-parameters that were used to train our models for reproduciblity purpose.

\begin{itemize}[topsep=0pt,itemsep=-1ex,partopsep=1ex,parsep=1ex]
    \item Number of epochs: $5$
    \item Learning rate: $5e-5$
    \item Batch size: $32$ (for models trained with an additional MLM objective: $16$ for each objective)
    \item Weight decay: 0
    \item Sequence length: $256$
\end{itemize}

Other settings were set to their default values (when using Huggingface's Trainer\footnote{\url{https://huggingface.co/docs/transformers/main_classes/trainer}} object).

\section{Annotation Guidelines}
\label{app:annotation guidelines}

\begin{figure*}
\centering
     \includegraphics[width=\textwidth, center]{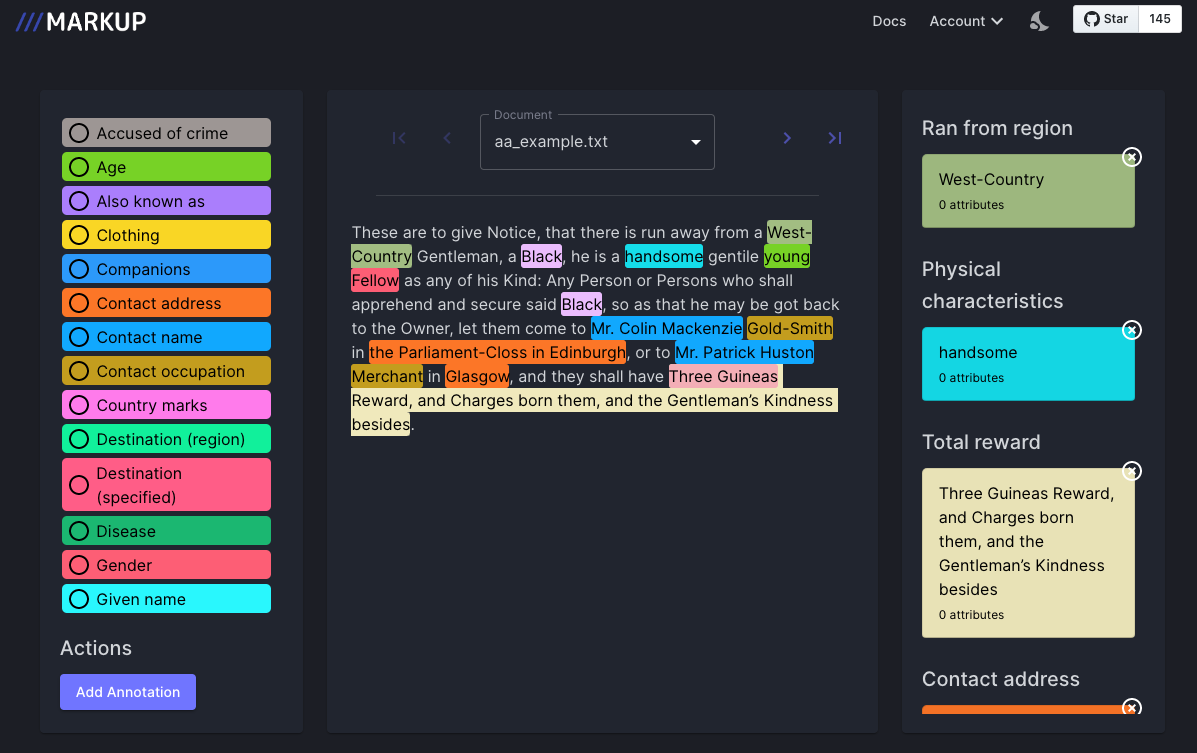}
     \caption{A screenshot of the annotation tool used by the experts. The ad shown here is an example that was presented to each expert, and they were instructed to annotate the other ads similarly.}
     \label{fig:annotation_tool}
\end{figure*}

Here we describe the annotation guidelines that were used for creating the evaluation set of the multilingual dataset. The experts were instructed to follow the same annotation scheme that was used to create the \textit{Runaway slaves in Britain} dataset. That is, given an ad, they were asked to find and mark in the ad the same 50 attributes that exist in the Runaway dataset (App. \ref{app:attributes}). More specifically, we asked the experts to familiarize themselves with the 50 attributes and ensured they understood them. We also supplied them with an English example to demonstrate how to perform the task and asked them to annotate the other ads in their respective language. To add an attribute, the annotators had to mark a span of text with their mouse and click on an attribute name from a color-coded list. Each attribute can be annotated more than once in each ad. Fig. \ref{fig:annotation_tool} shows a screenshot of the annotation tool that we used (Markup\footnote{\url{https://getmarkup.com/}}) and the English example.

\section{Additional Results}
\label{app:additional results}

\begin{figure}[t]
\centering
     \includegraphics[width=\columnwidth]{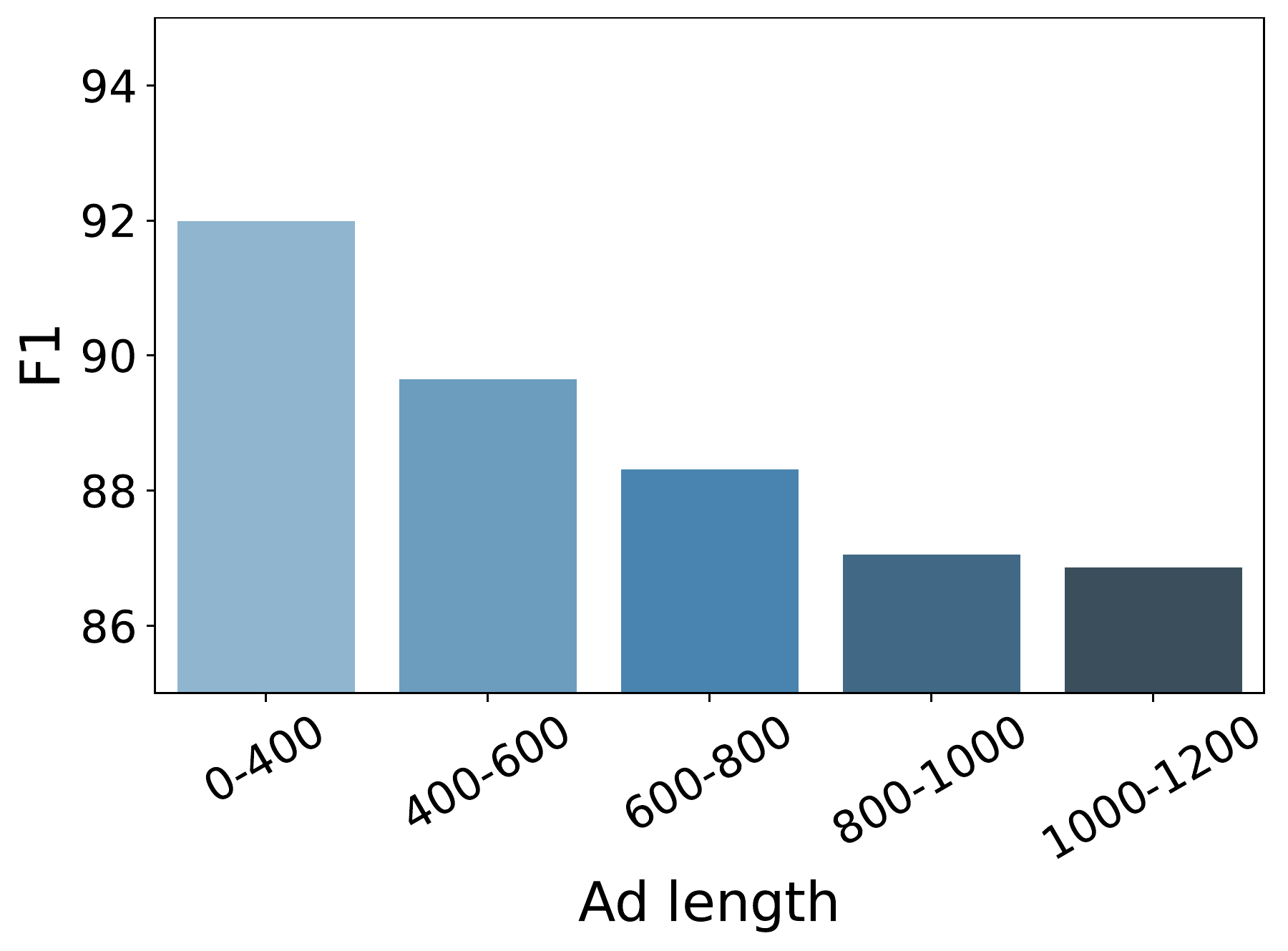}
     \caption{Performance of RoBERTa, fine-tuned on SQuAD-v2, on the English dataset. The longer the ad, the worse the model performs.}
     \label{fig:ad_length}
\end{figure}

\begin{figure}[t]
    \centering
    \begin{subfigure}{\columnwidth}
        \centering
        \includegraphics[width=\linewidth, center]{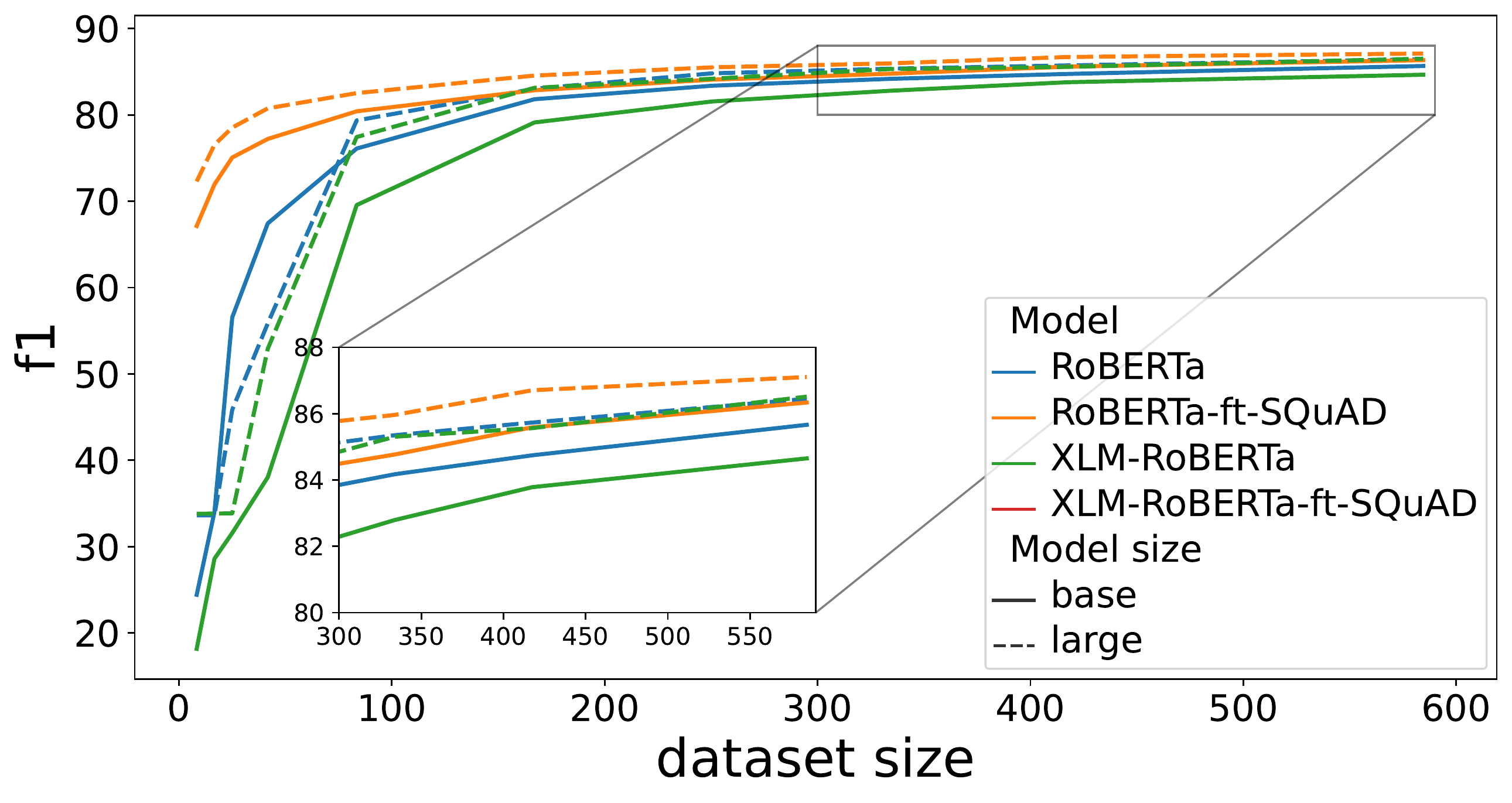}
        \caption{English}
        \label{fig:english_size}
      \end{subfigure}%
      
      \begin{subfigure}{\columnwidth}
        \centering
        \includegraphics[width=\linewidth, center]{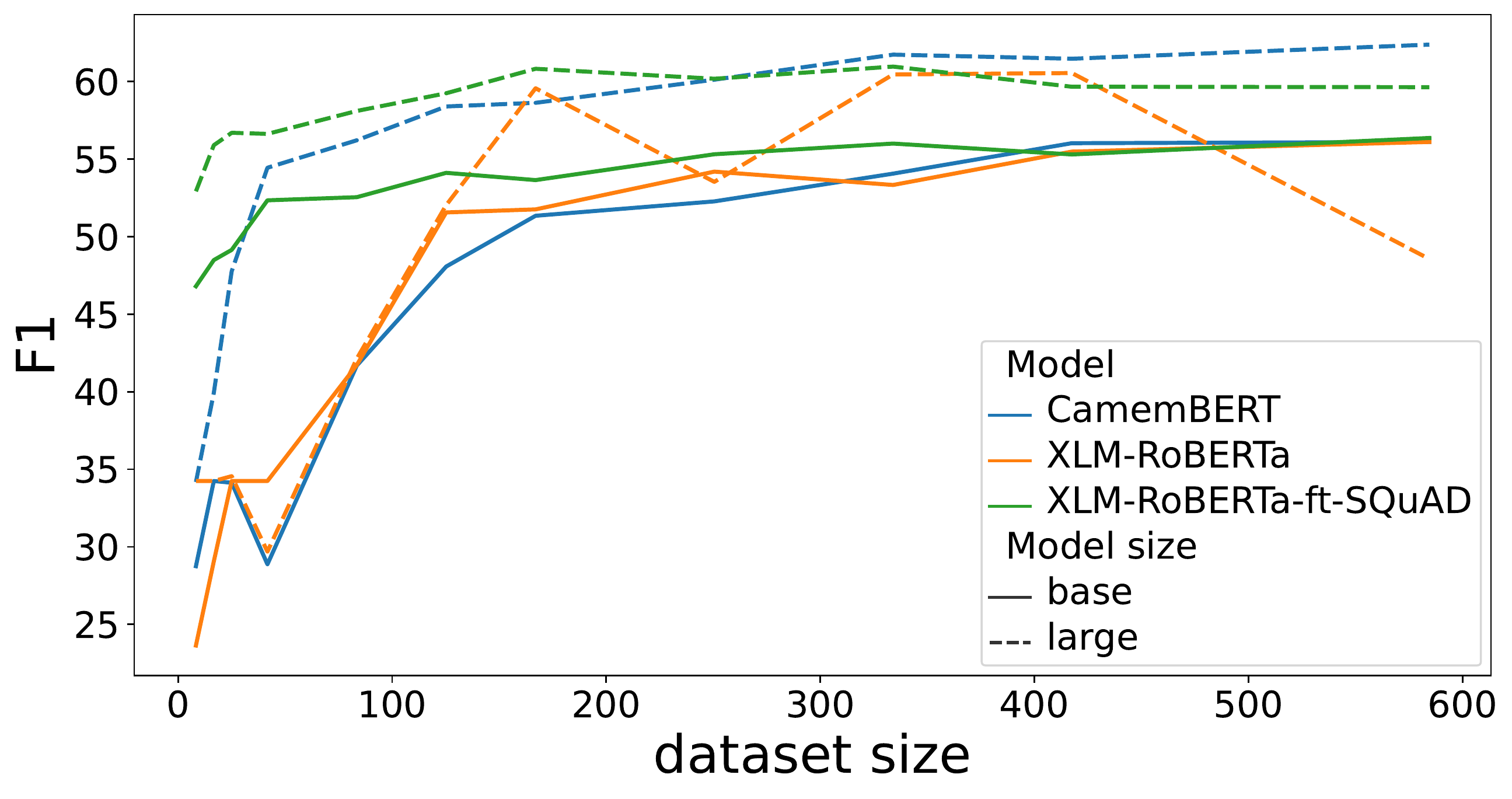}
        \caption{French}
        \label{fig:french_size}
      \end{subfigure}%
      
      \begin{subfigure}{\columnwidth}
        \centering
        \includegraphics[width=\linewidth, center]{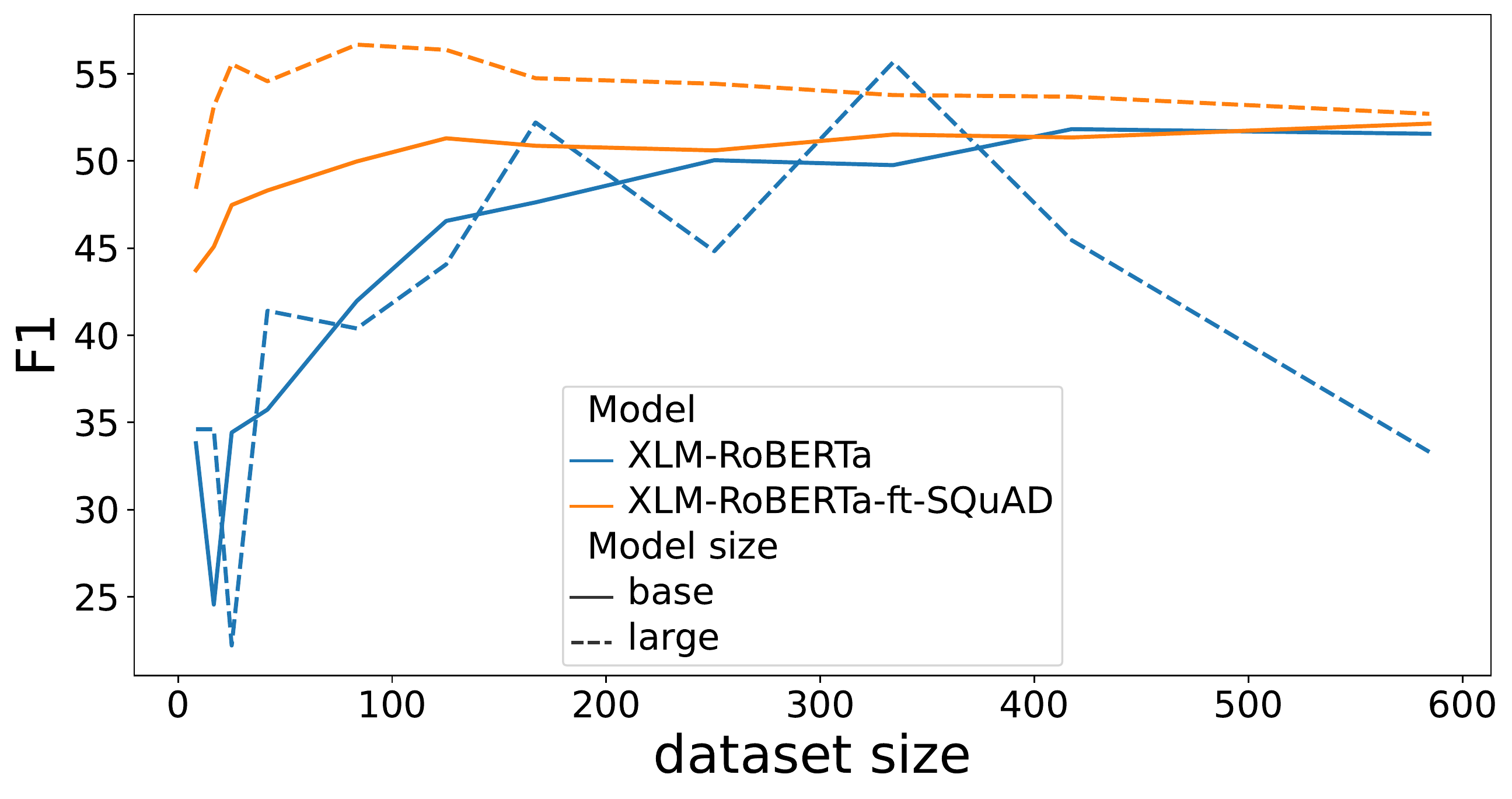}
        \caption{Dutch}
        \label{fig:dutch_size}
      \end{subfigure}

    \caption{Performance of models of different sizes on the Runaway dataset. The large models perform better than the base models for almost all cases in English, but tend to be more unstable in the other two languages. Unfortunately, not every model in French and Dutch is available in its larger version. Figures \ref{fig:french_size} and \ref{fig:dutch_size} include only the models for which both the base and the large version exist.}%
    \label{fig:model_sizes}%
\end{figure}

\begin{table*}
    \centering
    \begin{tabular}{cllcccc}
    \toprule
        \multirow{2}{*}{Language} & \multirow{2}{*}{Model} & \multirow{2}{*}{Setting} & \multicolumn{4}{c}{Dataset size} \\  \cmidrule{4-7}
         ~ & ~ & ~ & 8 & 16 & 25 & 585 \\ \midrule
         \multirow{8}{*}{en} & \multirow{4}{*}{RoBERTa-base} & None & 24.42 & 67.43 & 76.1 & 85.66 \\  
        ~ & ~ & further pre-trained & 15.22 & 69.52 & 77.59 & 85.85 \\ 
        ~ & ~ & MLM & 33.13 & 71.32 & 78.06 & 86.22 \\ 
        ~ & ~ & tri-training & 37.27 & 73.72 & 79.65 & 86.1 \\ \cmidrule{2-7}
        ~ & \multirow{4}{*}{RoBERTa-base-ft-SQuAD2} & None & 67.13 & 77.2 & 80.41 & 86.33 \\  
        ~ & ~ & further pre-trained & 57.18 & 76.52 & 79.93 & 85.91 \\ 
        ~ & ~ & MLM & 68.28 & 78.17 & 80.8 & 86.17 \\
        ~ & ~ & tri-training & \textbf{70.97} & \textbf{79.48} & \textbf{82.42} & \textbf{87.04} \\ \midrule
        \multirow{8}{*}{fr} & \multirow{4}{*}{CamemBERT-base} & None & 28.75 & 28.87 & 41.68 & 56.1 \\  
        ~ & ~ & further pre-trained & 26.33 & 24.13 & 40.82 & 57.93 \\ 
        ~ & ~ & MLM & 23.38 & 34.24 & 44.13 & 58.5 \\
        ~ & ~ & tri-training & 17.11 & 30.9 & 48.77 & 56.98 \\ \cmidrule{2-7}
        ~ & \multirow{4}{*}{CamemBERT-base-ft-SQuAD2} & None & \textbf{47.3} & \textbf{54.55} & 55.26 & 60.19 \\  
        ~ & ~ & further pre-trained & 34.04 & 49.48 & 54.04 & 61.01 \\ 
        ~ & ~ & MLM & 46.79 & 48.2 & 47.11 & 49.64 \\ 
        ~ & ~ & tri-training & 46.76 & 53.87 & \textbf{55.98} & \textbf{61.58} \\ \midrule
         \multirow{7}{*}{nl} & \multirow{3}{*}{RobBERT-base} & None & 34.61 & 34.61 & 35.76 & 48 \\ 
        ~ & ~ & further pre-trained  & 34.61 & 34.24 & 37.03 & 49.02 \\ 
        ~ & ~ & MLM & 42.84 & 43.29 & 43.67 & 46.35 \\ \cmidrule{2-7}
        ~ & \multirow{3}{*}{RobBERT-base-ft-SQuAD2} & None & \textbf{44.04} & 46.12 & 45.56 & 48.11 \\  
        ~ & ~ & further pre-trained & 34.61 & \textbf{46.16} & \textbf{48.15} & \textbf{49.84} \\ 
        ~ & ~ & MLM & 31.6 & 41.62 & 40.22 & 43.82 \\ 
        \bottomrule
    \end{tabular}
    \caption{$F1$ score of the models in semi-supervised settings. ``None'' means that no unannotated data were used. In ``further pre-trained'' we first further pre-train the model on an MLM objective and then fine-tune it on our annotated dataset. In ``MLM'' we train the models on an MLM and QA objective simultaneously. Finally, in ``tri-training'' we train the models using the tri-training algorithm. This line is missing from the Dutch models as the unlabeled Dutch dataset contains entire newspaper issues and not individual ads}
    \label{tab:semi_supervised_app}%
\end{table*}

\begin{table*}
    \centering
    \begin{tabular}{cllcccc}
    \toprule
        \multirow{2}{*}{Language} & \multirow{2}{*}{Model} & \multirow{2}{*}{Setting} & \multicolumn{4}{c}{Dataset size} \\  \cmidrule{4-7}
         ~ & ~ & ~ & 8 & 16 & 25 & 585 \\
        \midrule
        \multirow{8}{*}{fr} & CamemBERT-base & None & 28.75 & 34.24 & 34.13 & 56.1 \\\cmidrule{2-7}
        ~ & CamemBERT-base-ft-SQuAD-fr & None & \textbf{47.3} & 49.68 & 50.8 & \textbf{60.2} \\ \cmidrule{2-7}
        ~ & \multirow{3}{*}{XLM-RoBERTa-base} & None & 23.63 & 29.06 & 34.24 & 56.1 \\ 
        ~ & ~ & Simple & 22.17 & 23.98 & 29.19 & 54.73 \\ 
        ~ & ~ & MLM & 33.36 & 29.93 & 25.57 & 55.63 \\ \cmidrule{2-7} 
        ~ & \multirow{3}{*}{XLM-RoBERTa-base-ft-SQuAD-fr} & None & 46.8 & 48.48 & 49.14 & 56.36 \\
        ~ & ~ & Simple & 46.08 & \textbf{51.01} & \textbf{51.45} & 56.28 \\ 
        ~ & ~ & MLM & 47.0 & 48.36 & 48.34 & 53.98 \\ \midrule
        \multirow{8}{*}{nl}  & RobBERT-base & None & 34.62 & 34.62 & 34.62 & 48.0 \\ \cmidrule{2-7}
         & RobBERT-base-ft-SQuAD-nl & None & 44.05 & 44.4 & 45.0 & 48.11 \\ \cmidrule{2-7}
         & \multirow{3}{*}{XLM-RobBERT-base} & None & 33.8 & 24.55 & 34.42 & 51.56 \\ 
         & ~ & Simple & 17.23 & 26.3 & 33.15 & 44.45 \\ 
         & ~ & MLM & 37.66 & 45.21 & 45.76 & 46.31 \\\cmidrule{2-7}
         & \multirow{3}{*}{XLM-RobBERT-base-ft-SQuAD-nl} & None & 43.73 & 45.08 & \textbf{47.47} & \textbf{52.14} \\ 
         & ~ & Simple & 43.32 & 44.84 & 44.79 & 46.63 \\
         & ~ & MLM & \textbf{45.94} & \textbf{45.34} & 47.1 & 48.5 \\ 
        \bottomrule
    \end{tabular}
    \caption{$F1$ score of the models in different cross-lingual settings. ``None'' means that no cross-lingual training were used. ``Simple'' is standard cross-lingual training and ``MLM'' marks that the model was trained using an MLM-objective in addition to the standard QA loss.}%
    \label{tab:cross_lingual_app}%
\end{table*}

\section{Attributes}
\label{app:attributes}

Tab. \ref{tab:attributes} lists the different attributes that we wish to extract from the advertisements. The column ``Question'' describes the question that we feed the models in order to retrieve that attribute, and $\#$Annotated contains the number of occurrences of the attribute in the annotated dataset. 

\begin{table*}[ht]
    \centering
    \begin{tabular}{lll}
    \toprule
    Attribute & Question & $\#$Annotated \\ \midrule
    Accused of crime & What crimes did the person commit? & 107 \\
    Also known as & What other aliases does the person have? & 103 \\
    Clothing & What clothes did the person wear? & 656 \\
    Companions & What are the names of the person's friends? & 49 \\
    Contact address & Where does the contact person of the ad live? & 740 \\
    Contact occupation & What does the contact of the ad do for a living? & 278 \\
    Country marks & What country marks does the person have? & 63 \\
    Destination (region) & What is the destination region of the person? & 15 \\
    Destination (specified) & What is the name of the destination? & 118\\
    Disease & What kind of diseases does the person have? & 91 \\
    Given name & What is the given name of the person? & 693 \\
    Given surname & What is the last name of the person? & 196 \\
    Injuries & How was the person injured? & 63 \\
    Language & What are the communication skills of the person? & 319 \\
    Literacy & What is the literacy level of the person? & 8 \\
    Motivation & Why did the person escape his owner? & 4 \\
    Name of contact & Who is the contact person for the ad? & 678\\
    Origin & Where does the person originate from? & 28 \\
    Other reward & What other rewards were offered? & 382 \\
    Owner & Who is the owner of the person? & 395 \\
    Owner address & Where does the owner of the person live? & 270 \\
    Owner occupation & What does the owner of the person do for a living? & 78 \\
    Personality & What are the personality traits of the person? & 15 \\ 
    Physical characteristics & What are the physical characteristics of the person? & 568 \\
    Physical scars & What scars does the person have? & 131 \\
    Plantation marks & What plantation marks does the person have? & 23 \\
    Racial descriptor & What is the ethnicity of the person? & 807 \\
    Ran from region & What is the name of the region the person escaped from? & 3 \\
    Ran from specified & What is the name of the place the person escaped from? & 406 \\
    Religion & What is the religion of the person? & 13 \\
    Runaway date & What was the date of the event? & 15 \\
    Skills & What is the set of skills of the person? & 55 \\
    Specified occupation & What does the person do for a living? & 98 \\
    Stutters & Does the person stutter? & 22 \\
    Total reward & How much reward is offered? & 780 \\ \bottomrule
    \end{tabular}
    \caption{The attributes of the \textit{Runaways} dataset }
 \label{tab:attributes}
\end{table*}

\end{document}